\documentclass{article}

\usepackage{graphicx}
\usepackage{arxiv}
\usepackage{amsmath}
\usepackage[utf8]{inputenc} 
\DeclareMathOperator*{\argmin}{arg\,min}
\DeclareMathOperator*{\off}{off} 

\usepackage[T1]{fontenc}    
\usepackage{hyperref}       
\usepackage{url}            
\usepackage{booktabs}       
\usepackage{amsfonts}       
\usepackage[caption=false,font=footnotesize]{subfig}
\usepackage{nicefrac}       
\usepackage{microtype}      %
\usepackage{lipsum}

\title{A self-organising eigenspace map for time series clustering}

\author{
  Donya Rahmani\thanks{Corresponding author: Tel.: +0-44-2380525156} \\
  School of Mathematics\\ University of Southampton\\ Southampton, UK\\
  \texttt{d.rahmani@soton.ac.uk} \\
   \And
 Damien Fay \\
  Department of Analytics\\ Logicblox/Infor\\ Atlanta, GA\\
  \texttt{Damien.fay@infor.com} \\
  \And
  Jacek Brodzki\\
  School of Mathematics\\ University of Southampton\\ Southampton, UK\\
  \texttt{J.brodzki@soton.ac.uk} \\
}  

\begin{document}
\maketitle

\begin{abstract}
This paper presents a novel time series clustering method, the self-organising eigenspace map (SOEM), based on a generalisation of the well-known self-organising feature  map (SOFM). The SOEM operates on the eigenspaces of the embedded covariance structures of time series which are related directly to modes in those time series. Approximate joint diagonalisation acts as a pseudo-metric across these spaces allowing us to generalise the SOFM to a neural network with matrix input. The technique is empirically validated against three sets of experiments; univariate and multivariate time series clustering, and application to (clustered) multi-variate time series forecasting. Results indicate that the technique performs a valid topologically ordered clustering of the time series. The clustering is superior in comparison to standard benchmarks when the data is non-aligned, gives the best clustering stage for when used in forecasting, and can be used with partial/non-overlapping time series, multivariate clustering and produces a topological representation of the time series objects. 
\end{abstract}


\section{Introduction}
Clustering of time series is an important and difficult problem that has drawn much attention recently and which in broad terms arises in situations where a set of unlabelled time series needs to be split into classes according to an appropriate 
notion of similarity. Among the many examples, such problems arise  in areas as diverse
as biology \cite{ernst2005clustering}, EEG clustering \cite{orhan2011eeg}, identification of planets from luminosity readings \cite{jiang2006data}, financial time series \cite{dias2015clustering}, synchronisation between economies \cite{papageorgiou2010business}, similarity between sensor data in the wild \cite{mittal2010wireless}, identification of seasonal and non-seasonal items in retail~\cite{AbhayJha2015}, to mention but a few. As noted in a survey by Liao~\cite{liao2005clustering} there are three fundamental approaches, i) those derived from the {\it{raw data}} (for example, correlation and warped distance measures), ii) {\it{features derived}} from the time series (for example, the variance), and iii) comparisons based on {\it{models}} of the time series (typically the parameters or residuals of those models). All of these approaches have in common that a vector of features (which may be just the raw data or the model parameters) are used as input to a similarity metric  followed by a clustering step using one of many standard clustering algorithms. 

In this paper we introduce the Self-Organising Eigenvector Map (SOEM) which is a novel approach to the clustering of time series. The SOEM does not fit neatly into any of the three categories of Liao, mentioned above. This method is a generalisation of the 
the Self-Organising Feature Map (SOFM or Kohonen map \cite{kohonen1982self}). 
In contrast to the first category of  methods mentioned above, 
the SOEM does not require the time series to be in some sense aligned or semi-aligned, by which we mean that a warping of time in one time series leads to a high correlation with another (See~\cite{guillame2017} for a detailed discussion of alignment). Indeed, the SOEM is ideal for time series which may appear distinct but evolve according to a similar underlying state space. In contrast to the second approach in Liao's classification (feature derived), the features derived from the time series are based on an embedding that does not require feature engineering  based on domain-specific prior knowledge. The SOEM shares some similarities with model-based methods of (iii), but 
it makes no  assumptions about the distribution or stationarity of the time series and does not require parameter tuning. 

As with any clustering technique, the success of the SOEM rests on the relevance of the {\it{targeted characteristic}} to the underlying class one wishes to uncover. The SOEM specifically targets the spectrum of the covariance structure of time series embeddings which is a rich structure that can be directly related to modes in time series (Section~\ref{s:ssa}).  It is proposed that the SOEM will find a place in the toolbox of the data analyst, and will be  particularly useful for non-aligned time series and time series which have different lengths but which evolve according to similar dynamics. 
 
Our approach draws together several well-known techniques from disparate fields. The starting point is the analysis of covariance matrices obtained through multivariate Singular Spectrum Analysis (MSSA). See 
\cite{Golyandina2001} for a description of the long  line of research into Singular Spectral Analysis (SSA)  of time series, and  \cite{Golyandina2005} for MSSA. In this field a key insight (for us a justification) is that sets of time series benefit from multivariate modelling (MSSA) if their time embedded covariance matrices share a common eigenstructure (see Section~\ref{s:ssa}). In particular, we demonstrate that clustering of time series according to the common eigenstructure of their covariance 
matrices  provides an efficient new clustering method for time series. The challenge of deriving common eigenstructures has long been part of another field of research, blind source separation  \cite{GLASHOFF20132503}, in which \emph{approximate JD} has emerged as 
a very useful and flexible methodology. We employ JD as a pseudo-semi metric between matrices making possible an iterative update scheme and thus a training algorithm. Finally the clustering algorithm itself brings together all the above elements in a SOFM variant which we call the SOEM. The SOEM clusters time series based on the covariance matrices of each time series as input with each node representing a basis; the update scheme utilizes JD. Note that the SOEM is a generalisation of the SOFM in the sense that it clusters {\it{matrices}} (as opposed to {\it{vectors}}) and to the best of our knowledge is the first SOFM variant to do so. Empirical results show that the algorithm does indeed cluster, performs well (especially for non-aligned time series), and performs topological ordering (a validation criteria for an SOFM~\cite{tatoian2016self}).    

The paper is structured as follows. In Section~\ref{s:related} we review relevant time series clustering techniques and contrast them with the SOEM. MSSA is not a central part of our investigation, but we provide a review in  Section~\ref{s:ssa} of SSA and in particular MSSA to provide a motivation and justification for the SOEM methodology. Section~\ref{s:background} presents Joint Diagonalization and considers its use as a metric. This leads onto the development of the SOEM model. Section~\ref{s:results} presents empirical results in 3 different real-world scenarios. The first seeks to cluster the (pre-labelled) UCR data sets from which we find that the SOEM detects clusters that are comparable to those found via the widely used tSNE algorithm while performing better when the data is less aligned.  The second scenario demonstrates how to extend the technique to a multivariate time series data set. The final scenario is used to examine forecasting performance where the forecasts are dependent on the clusters derived. In that Section we also examine the topological ordering resulting from a SOEM map showing that it does indeed represent a metric space.  
\section{Related Work}\label{s:related}
In this section we give a brief overview of time series clustering techniques and focus on alignment (or lack thereof) and features derived from the time series. We also provide a discussion of research related to the core technique of interest in this paper: the self-organising feature map.

Time series clustering generally involves using time series representations in order to reduce their dimensionality. Proposed representations include the Discrete Fourier Transform \cite{agrawal1993efficient}, Discrete Wavelets Transform \cite{chan1999efficient},
Singular Value Decomposition \cite{keogh2001locally}, Adaptive Piecewise Constant Approximation \cite{keogh2001locally}, and 
Symbolic representations \cite{lin2007experiencing} etc. 
The reduced representations or features can be then used to compute clustering of the time series. Almost all clustering techniques require a measure to compute the distance or similarity between the
time series being compared.
Most commonly used distance or similarity measures used in the literature (\cite{liao2005clustering,keogh2003need,rani2012recent,aghabozorgi2015time}) are Euclidean distance, Dynamic Time Warping (DTW), distance based on Longest Common Sub-sequence, Sequence Weighted Alignment model, Edit Distance on Real sequences, Spatial Assembling Distance, etc.

However, these measures assume that the time series are \emph{aligned} or semi-aligned, i.e. time series in the same clusters evolve closely at the same or at similar times, see \cite{aghabozorgi2015time,rani2012recent,keogh2001locally}.
For example, DTW seeks  the optimal alignment between two time series by warping time (adjusting the time index dynamically to obtain a good match) and then, for instance, calculate the Euclidean distance between them \cite{xu2015dynamic}. The optimal alignment is achieved by mapping the time axis of one time series onto the time 
axis of the other one,  which is basically the same as the optimal warping function. See \cite{sakoe1978dynamic} for a thorough description on DTW. In the literature, there are several variants of the warping function, such as Weighted Dynamic Time Warping (WDTW), Derivative Dynamic Time Warping and multiscale DTW specifically for speeding up the DTW (we refer to \cite{muller2007dynamic} for further explanation of the techniques mentioned above).  However, if the time series are not semi-aligned, then DTW by itself fails to cluster time series accurately. As an example, Figure~\ref{f:lrf_mess} shows  three time series with the same dynamics which are obviously not aligned 
and even semi-aligned. 

In classification, non-aligned time series are often classified by different means such as kernel methods or motif extraction. In essence the algorithms seek some characteristic of the data which is not localised to a particular time. \cite{jeong2015support} proposed a new kernel function based on a WDTW distance for multiclass support vector machines for \emph{classifying} non-aligned time series. This provides an optimal match between two time series by not only allowing a non-linear mapping between two sequences, but also considering relative significance depending on the phase difference between points on time series data. Clustering time series under time warp measures is very challenging and as yet unresolved, as it requires aligning multiple temporal series simultaneously given an unlabelled time series \cite{soheily2016generalized}.

In a recent paper Keogh and Kasetty~\cite{keogh2003need} re-examined the performance of multiple clustering algorithms over multiple benchmarking data sets. They conclude that most existing clustering techniques do not work well due to the complexity of their underlying structure and data dependency. This indeed causes a real challenge in clustering temporal data of  high dimensionality, unequal length, complicated temporal correlation, and a substantial amount of noise \cite{soheily2016generalized}. For example, the SOFM, which is an unsupervised learning algorithm, does not work well with
time series of unequal length due to the difficulty involved
in defining the dimension of weight vectors \cite{liao2005clustering}. In summary, the choice of the specific clustering algorithm can be highly problem dependent and often real-world data sets do not adhere to the underlying assumptions of the algorithm. 

We propose here a new approach to clustering of time series based 
on our self-organising map for temporal data which are described by matrices rather than feature vectors. This provides and advances over the existing methodology.  A predecessor to this work  \cite{seo2004self} clusters pairwise data and co-occurrence data using a variant of the SOFM.  However this approach is based on a single matrix derived from the data as opposed to clustering multiple matrices as presented here. To the best of our knowledge the proposed technique is the first to present a truly matrix-based generalisation of the SOFM. In addition, our technique is applicable to non-aligned time series, time series of differing lengths, multivariate time series clustering and requires minimal tuning. 
\section{Background on MSSA}\label{s:ssa}
In this Section we review SSA and MSSA in as far as it provides foundations 
and justification for the SOEM. SSA is a technique for modelling, forecasting and reconstructing time series which uses a time delay embedding and the Singular Value Decomposition (SVD). The technique has been widely used for modelling EEG signals~\cite{Kouchaki2015}, financial times series~\cite{Thomakos2002}, change point detection~\cite{Moskvina2003}, and medical times series~\cite{Sanei2011} to mention but a few.

\begin{figure}
    \centering
    \includegraphics[width=0.8\textwidth]{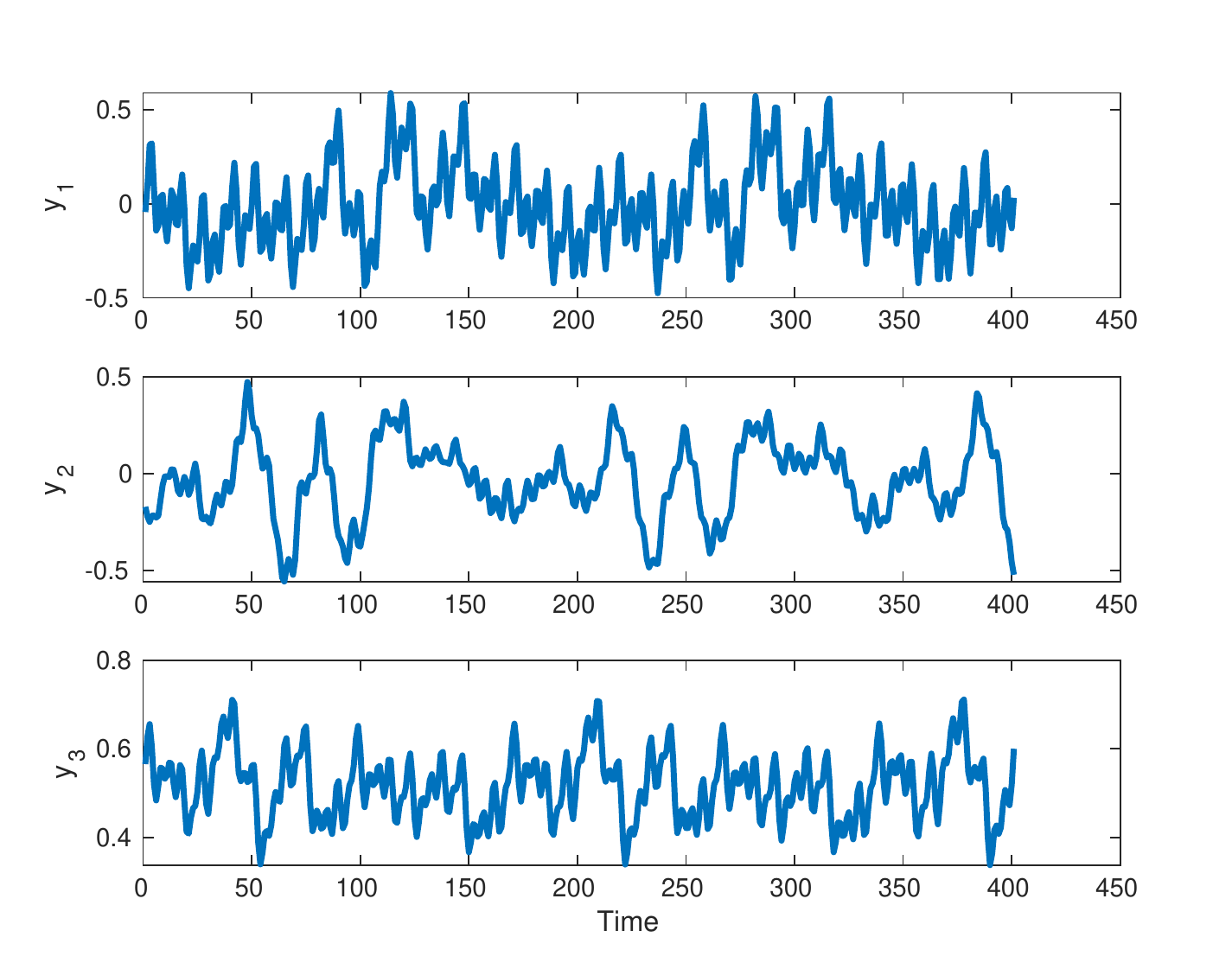}
    \caption{Three time series which share a common LRF.}
    \label{f:lrf_mess}
\end{figure}

The first step in SSA involves embedding a time series into what is called the trajectory matrix, $\mathbf{X}$, as: 
\begin{equation} \label{e:traj}
\mathbf{X} ={ \begin{pmatrix} 
y_{1} & y_{2} & \cdots & y_{N-L+1} \\
y_{2} & y_{3} & \cdots & y_{N-L+2} \\
\vdots & \vdots & \ddots & \vdots \\
y_{L} & y_{L+1} & \cdots & y_{N}
\end{pmatrix}
}\\  \end{equation}
where $y_t$ is the time series at time $t$ of length $N$, $L$ is called the embedding dimension and is the maximum delay applied to the time series to form the embedding. We explain below how to select the embedding dimension.  The next step involves estimating a lower dimensional reconstruction of $\mathbf{X}$ as: 
\begin{equation}\label{e:recon} 
{\widehat{\mathbf{X}}} = \sum_{j=1}^{r}  U_j U_j^T \mathbf{X}
\end{equation}
where $U_j$ is the eigenvector corresponding to the $j^{th}$ eigenvalue of $\mathbf{X} \mathbf{X}^T $ obtained using the SVD, and $r$ is the number of retained components. In some applications it is preferable to retain a subset of the eigenvectors rather than those ranked by their algebraic singular values. A lower dimensional representation of the time series may then be constructed via averaging across the  anti-diagonals of $\widehat{\mathbf{X}}$ which decompose the time series into a trend and seasonal components (see~\cite{Golyandina2001} for more details). From Equation~(\ref{e:traj}) note that should $y_{N+1}$ become available this would result in a new column appended to the trajectory matrix with the first $L-1$ elements of the column being (known) $[y_{N-L+2} \dots y_{N}]^T$. Indeed this may be used to produce a forecast by noting that the first $L-1$ rows of $\widehat{\mathbf{X}}$ can form a linear projection onto the $L^{th}$ row as: 
\begin{equation}
 \hat{y}_{N+1}=\sum_{i=1}^{L-1} \phi_i y_{N-L+1+i}
\end{equation}\label{LRF}
where $\Phi = \{\phi_1 \ldots \phi_{L-1}\}$ are the coefficients of projection known as the \textit{linear recurrent coefficients} (LRF). We may now discuss the selection of the embedding dimension $L$. It should be large enough to capture the dynamics of the times series (for example, if the time series is seasonal, $L$ should be greater than the seasonal length). However there is a trade-off as for a larger $L$ we capture more dynamics, but there are fewer columns in $\mathbf{X}$ with which to estimate the dynamic coefficients which also have a larger dimension (see Equation~(\ref{e:lrf})). 
The LRF coefficients may be estimated from the eigenvectors of $\widehat{\mathbf{X}}$ as follows:
\begin{equation}\label{e:lrf}
\Phi=\frac{1}{1-\sum\limits_{i=1}^{r} \pi^2_{j}}\sum_{j=1}^{r}\pi_{j}U_{j}^{\bigtriangledown}
\end{equation}
where the eigenvectors of $\widehat{\mathbf{X}}$, $U_j$, have been partitioned into the first $L-1$ rows, $U_{j}^{\bigtriangledown}$, and the last element, $\pi_{j}$, i.e.  $U_j  = [U_{j}^{\bigtriangledown} \pi_{j}]^T$. Equation (\ref{e:lrf}) has many interesting properties (see~\cite{Golyandina2001} for a discussion and proof); it may be used with the chain rule of forecasting to provide a multi-step ahead forecast. Secondly, it is not a typical AR(L-1) model as the residual is not i.i.d.\footnote{A trivial example has $r=1$ which would mean the residual has seasonal components which are auto-correlated i.e. not i.i.d.}, indeed $\Phi$ encapsulates the \textit{dynamics} of the time series (specifically it defines the \textit{linear recurrence}~\cite{Golyandina2001})  and gives an optimal multi-step ahead forecast while an AR model typically gives the optimal 1-step ahead forecast (this is a major advantage of SSA as a forecasting tool).  Critically, for our purposes, note that $\Phi$ is \textit{determined solely by the eigenvectors} $U_j$. Experimenting with different LRF coefficients can give one a sense of the power of SSA and in particular the huge variety of time series that can be represented; for example Figure~\ref{f:lrf_mess} shows three sample paths with very different characteristics. However, they have all been generated from the same LRF, and, by extension share the same covariance structure. 

The Multivariate version of SSA (MSSA), follows the procedure above but given a \textit{set} of $M$ time series, forms a stacked trajectory matrix $\widehat{\mathbf{X}}_H$ as: 
\begin{equation}\label{e:mssa}
\widehat{\mathbf{X}} _H =  [\widehat{\mathbf{X}} ^{(1)} \ldots \widehat{\mathbf{X}}^{(M)}] 
\end{equation}
where $\mathbf{X}^{(i)}$ is the trajectory matrix for the $i^{th}$ time series. MSSA then proceeds by performing the SVD on  $\widehat{\mathbf{X}}_H \widehat{\mathbf{X}}_H ^T$ (several variants exist see~\cite{Golyandina2001}). As pointed out in~\cite{Golyandina2005} MSSA will outperform (multiple) univariate SSA if two (or more) time series share a \textit{common eigenstructure}. This is called component matching and has been discussed in several papers (see~\cite{Rodrigues2018},\cite{Golyandina2010} and references therein). In the real-world, however, exact component matching rarely occurs~\cite{Golyandina2005} and one seeks time series with \textit{close} eigenstructures. One measure of closeness is the \textit{F-feedback} test statistic of~\cite{Hassani2010f} using the same principle as Granger causality; this is limited in that one requires ${M^2}$ tests for bivariate matching, $M!$ for a full multivariate comparison. 
Furthermore, because of a lack of symmetry as $F(A,B) \ne F(B,A)$, $F$ is not a metric and so cannot be used to give a lower dimensional representation of how the time series relate to each other. The key point is that the similarity between the eigenvectors of the covariance matrices reflects similarity of the underlying modes driving the dynamics in the corresponding time series. 
\section{Joint Diagonalisation and the SOEM}\label{s:background}
In this section we develop the proposed algorithm. The first building block is joint diagonalisation which forms the metric used in the update for the SOEM. 

\subsection{Joint Diagonalisation}\label{s:jd}
As is well-known, if two square matrices commute, they share a common eigenvector, and there exist various procedures for joint diagonalisation of 
a set of commuting matrices; see, for example, 
\cite{cardoso1996jacobi}. However, for a set $\textbf{\it{C}}$ of real symmetric matrices which do not commute,  one may seek an approximate solution to the problem of joint diagonalisation. An interesting idea, proposed in 
the closing remarks of \cite{cardoso1996jacobi}, is to construct an '\textit{average}' set of eigenvectors via approximate Joint Diagonalisation (JD). For a 
square matrix $A$ we denote by $\off(A) = \sum_{1\le i\ne j \le N} {a^2_{i,j}}$, and we also define the sum of off-diagonal elements as $\delta(\textbf{\it{C}}) = \sum_{m=1}^{M} \off(UC_m U^T)$.
Then approximate JD is a procedure that 
for a set of non-commuting symmetric matrices 
$\textbf{\it{C}} = [C_1 \ldots C_M]$ returns a unitary 
matrix $\overline{U}$, the columns of  which are regarded as a set of approximate common eigevectors for the matrices in $\textbf{\it{C}}$. The matrix $\overline{U}$ is given by
\begin{equation}\label{koh_eqn}
\overline{U} = \argmin_{U} \sum_{m=1}^{M} \off(UC_m U^T) 
\end{equation} 
where the minimum is taken over the set of unitary matrices $U$. In practice, this is achieved via an iterative algorithm in which, at each step, a Jacobi rotation is selected to optimise a joint diagonality criterion (see ~\cite{cardoso1996jacobi} for details). Another important feature of approximate JD is that it allows one to  define a pseudo-semi metric\footnote{A pseudo metric is one for which the axiom of identity does not hold for some values, a semi metric is one for which the triangle inequality does not always hold.} on the space of symmetric matrices as follows (see ~\cite{GLASHOFF20132503} for a similar discussion). It is straightforward to see that $\delta([C_i,C_j])$ is non-negative $\delta([C_i,C_j])\ge 0$ (by virtue of the quadratic sum), and symmetric $\delta([C_i,C_j])  = \delta([C_j,C_i])$ (as the matrix order is interchangeable). Note that it is possible to have $\delta([C_i,C_j]) = 0$ while  $C_i \ne C_j$; this happens for instance when $C_i$ and $C_j$ share the same eigenvectors but with different eigenvalues. However, as will be seen in Section~\ref{s:soem} we will use $C_i = \mathbf{X}^{(i)} {\mathbf{X}^{(i)}} ^T$ and as pointed out in Section~\ref{s:ssa}, the LRF components depend only on the eigenvectors; i.e. we may classify them as equivalent. The triangle inequality is interesting. As shown in~\cite{GLASHOFF20132503} a simple example may be constructed for which $\delta([C_i,C_j])  > \delta([C_i,C_k]) + \delta([C_k,C_j])$ proving that the triangle inequality does not hold in general. However, they show that as the dimension of $C_i$ increases the probability of violating the triangle inequality tends asymptotically to zero and show empirically that the probability falls rapidly with $N$.  To illustrate this point, we tested 250K triplets from the SOEM input matrices in the UCR datasets (Section~\ref{s:ucr_results}) and did not find a single offending case.

\subsection{The SOEM}\label{s:soem}
We  now consider a set of time series and their associated trajectory matrix covariances  $\mathbf{X} \mathbf{X}^T$ (Section~\ref{s:ssa}) which are symmetric, positive definite real matrices. We also assume given the pseudo-semi metric 
$\delta$  (Section~\ref{s:jd}), which reflects the ability 
of the time series to reinforce one another. Our clustering 
procedure relying on the SOEM takes as input a set of real positive definite matrices and assigns each to a cluster defined by an orthornormal matrix using competitive learning with respect to the JD pseudo-metric $\delta$. In this we 
depart from the standard self-organising feature map (SOFM)
methodology which takes as input a set of vectors and assigns each to a cluster defined by a vector of weights. Training is performed using competitive learning with the metric defined as the Euclidean distance between the vector input and vector weights.

To explain our new method, let us recall the basics of 
the SOFM. An SOFM consists of a grid of topologically distributed nodes each with an associated function. An input is presented to this grid (i.e. the functions are evaluated for the input at every node) and the node which activates with the minimum distance 
is called the \emph{winner}. During training the function of winning nodes and, crucially, the functions of its neighbours are updated so as to reinforce the activation. Assuming the network training converges, the distribution of the winning nodes reflects the topological ordering of the inputs, in addition, the nodes should place themselves so that the input which maximises them is sampled from the same underlying distribution as that of the input space~\cite{tatoian2016self}. In such a case the 2-D grid of nodes may be considered a lower dimensional mapping of the input space distribution and as such is useful for clustering. Bearing this in mind, the SOEM procedure
consists of the following steps. The structure of this 
procedure has the same general shape as the Kohonen map 
SOFM, but where each step has been 
generalised to the case of matrices. 

\begin{enumerate}
\item {\bf The competitive step}. 
Each node has an associated orthonormal basis, $\{U_{i,j}\}$ where $i,j$ denote the grid locations. Given a set of input matrices $\textbf{\it{C}} = [C_1 \ldots C_M]$ the winning node, $s$, is that which aligns best with the projection space of the matrix input, $C$, as: 
\begin{equation}\label{e:soem_opt}
s = \argmin_{i,j} \normalfont{\off}_2 \{ (U_{i,j}^TU_{i,j})^{-1}U_{i,j}^T C U_{i,j}\}
\end{equation}
\item {\bf The update step}. 
The node bases are then rotated towards the input matrices according to: 
\begingroup\makeatletter\def\f@size{8}\check@mathfonts
\def\maketag@@@#1{\hbox{\m@th\large\normalfont#1}}
\begin{equation} 
U_{i,j} \leftarrow  \mathbf{J}_2 \Big( \vec{v}_{i,j}U_{i,j},  h(s_1,i,j)C_1  ,\ldots, h(s_M,i,j)C_M \Big)
\end{equation}\label{e:soem_update}
\endgroup
where $\mathbf{J}_2 (\cdot)$ is the joint diagonalisation function which allows simultaneous diagonalisation of the old eigenvectors set $U_{i,j}$ with a weighted set of the input matrices, $\nu$ is a gain term $\in[0,1]$ and $h(s_k,i,j)$ is a monotonically decreasing function with the distance from node $\{i,j\}$ to $s_k$. Note that for an SOFM each input is presented one by one, thus updates occur iteratively, while for an SOEM all the inputs matrices are used simultaneously to update a basis\footnote{We experimented with iterative updates but found that the eigenspaces are adapted either too slowly or quickly to each input making convergence difficult to achieve; however, we did not exhaustively examine possible solutions.}.
  
\item {\bf The iteration step}. After all inputs have been presented the algorithm typically adapts. For example, the input value for $\nu$ may be reduced which results in slowing the refinement of the node vectors to stop oscillations. Also, $h(s_k,i,j)$ may be changed, for example if $h(s_k,i,j) = \mathcal{N}(0,\sigma_h)$ is a Gaussian kernel function then the width of this kernel may be reduced after each iteration thus allowing a local rather than global refinement of the solution.
\end{enumerate}

Following step 3 the algorithm returns to step 1 and iterates until a stopping condition is found. In this study a set number of iterations is given although the algorithm may be stopped, for example, when the kernel's effective radius falls below one. 
The SOEM is initialised by assigning random vectors to each node which are then adjusted to form a basis using Gram Schmidt orthogonalization. In addition, each input is scaled so that it has unit norm. In the first iteration, the winning nodes are assigned randomly; this ensures the first iteration spans the entire eigenspace of the inputs. In the following section we examine the SOEM's performance on real-world data sets.

\section{Results}\label{s:results}
In this section, we examine the ability of the SOEM to perform clustering using 3 different real-world data sets. The first set is the well-known UCR time series repository~\cite{UCRArchive}. We used the first 
27 time series from this collection of 85 inputs. \footnote{This was to reduce the computational burden and the volume of results.} Each of the time series is labelled and the data are typically \textit{aligned} (see \cite{guillame2017}). The second data set is taken from the UAE multivariate time series archive~\cite{bagnall2018uea}, the articulary word recognition data set.
The aim is to evaluate SOEM performance on a multivariate time series data set. 
The final data set is the unemployment rates of the USA states. This is unlabelled, semi-aligned and geographically distributed (i.e. neighbouring states are known to be similar to some degree). We examine the ability of the SOEM to identify clusters seeking validation via subsequent MSSA forecasting. In addition, we validate the topographic accuracy of the technique. Note that our focus 
is clustering and not classification, so there all the data is used for training with no test set (except for the last section where the SOEM is a pre-processing step). 

The SOEMs trained below all consist of a 2-D grid of size $30 \times 30$, the grids are bounded (i.e. nodes on the edge do not neighbour opposite nodes). The number of iterations is set to 10. A Gaussian kernel, initially with an effective radius $1/4$ the size of the grid is employed, i.e. $\sigma_h = 30/4$. This allows each input to affect a large section of the grid. After each iteration the kernel is reduced in size as $\sigma_h = 1/(4+i)$, where $i$ is the iteration number and at the last iteration this covers only direct neighbours of the winning nodes. There are many different ways to configure a Kohonen network, the values used above gave reasonable performance but other configurations may be superior.

\subsection{UCR datasets}\label{s:ucr_results}
The first analysis is conducted across 27 UCR time-series data sets. As there are many time series in this collection the embedding dimension is set to $L = \lceil N/10 \rceil$. Typically an embedding dimension is set by hand after close examination of a time series to determine that the embedding covers the full evolution of the time series' dynamics. As a comparison benchmark we use the popular t-Distributed Stochastic Neighbour Embedding (t-SNE) \cite{maaten2008} using projections into 2 dimensions for fair comparison. As we have given class labels for the data sets we can measure the scatter of the locations (t-SNE) or grid points (SOEM) for each class and we can also measure the separation of the classes. Specifically, we use the Davies-Bouldin (DB) index~\cite{davies1979}. The DB index takes the average similarity between each class and its most similar one, then averages over all the classes. Therefore, the best clustering scheme essentially minimises the DB index. 

\begin{figure*}
        \centering
            \subfloat [\label{f:it_ex_1}]{%
          \includegraphics[width=0.325\textwidth]{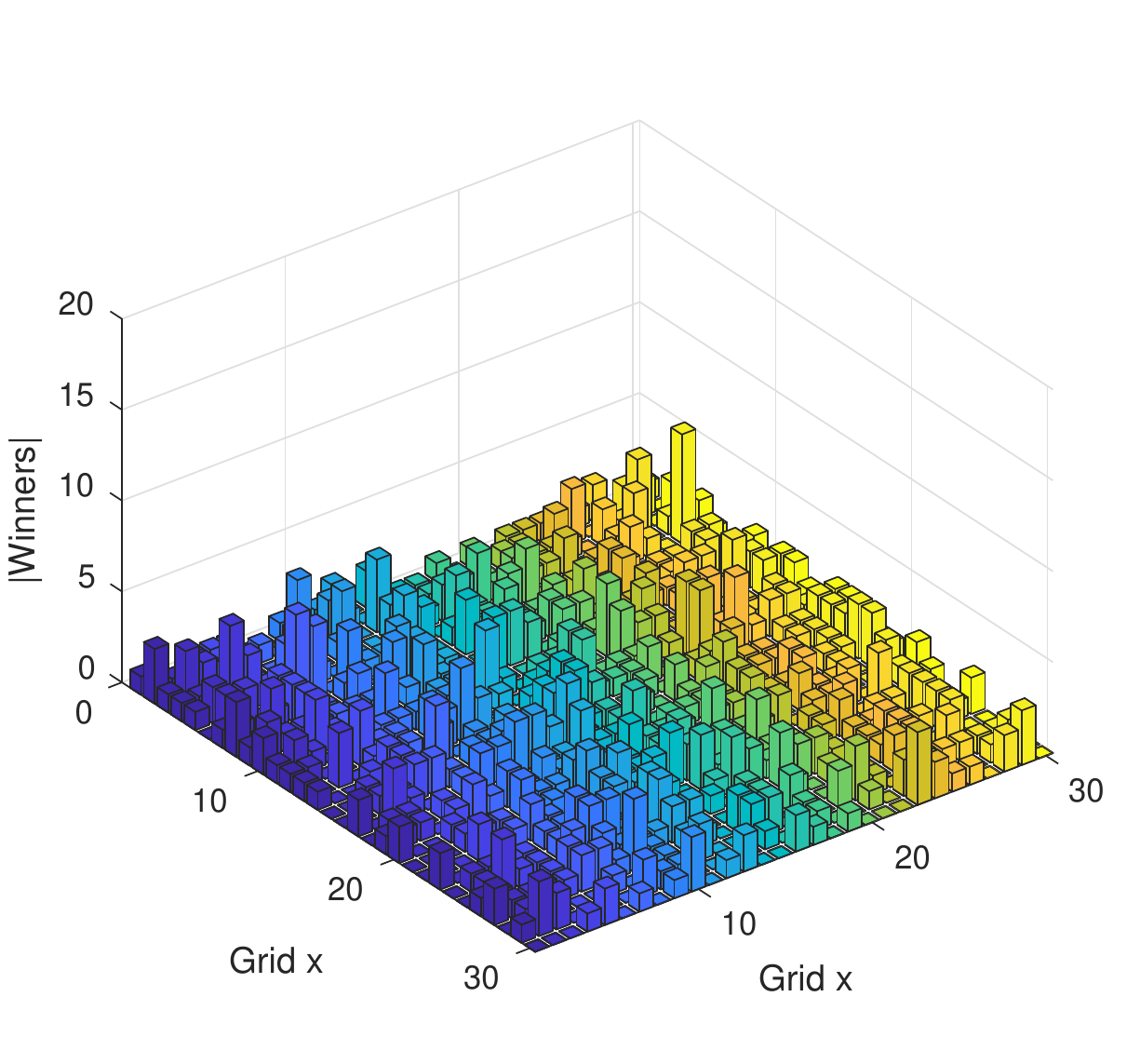}                   }
             \subfloat[\label{f:it_ex_2}]{%
          \includegraphics[width=0.325\textwidth]{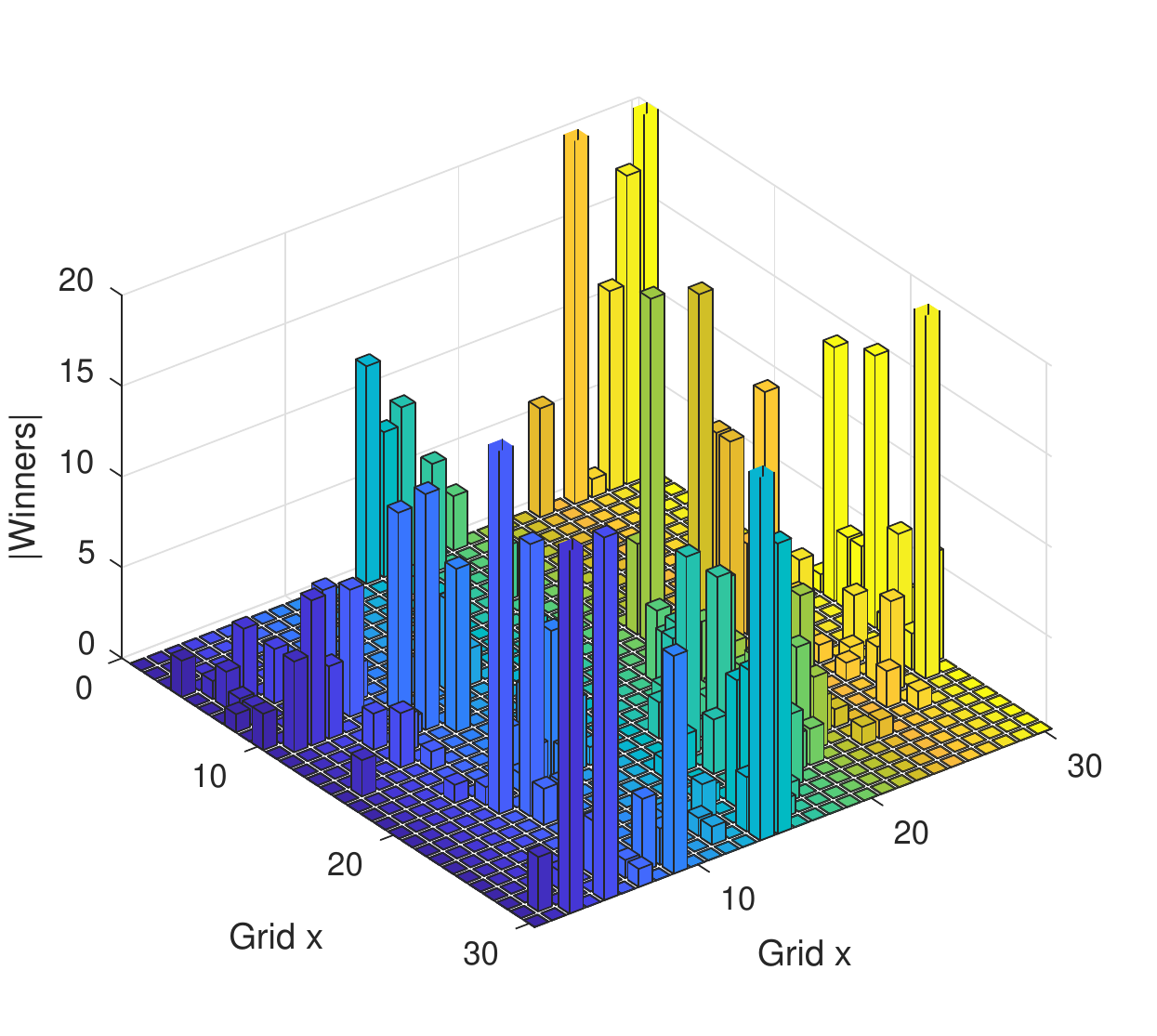}                   }
            \subfloat[\label{f:it_ex_3}]{%
          \includegraphics[width=0.325\textwidth]{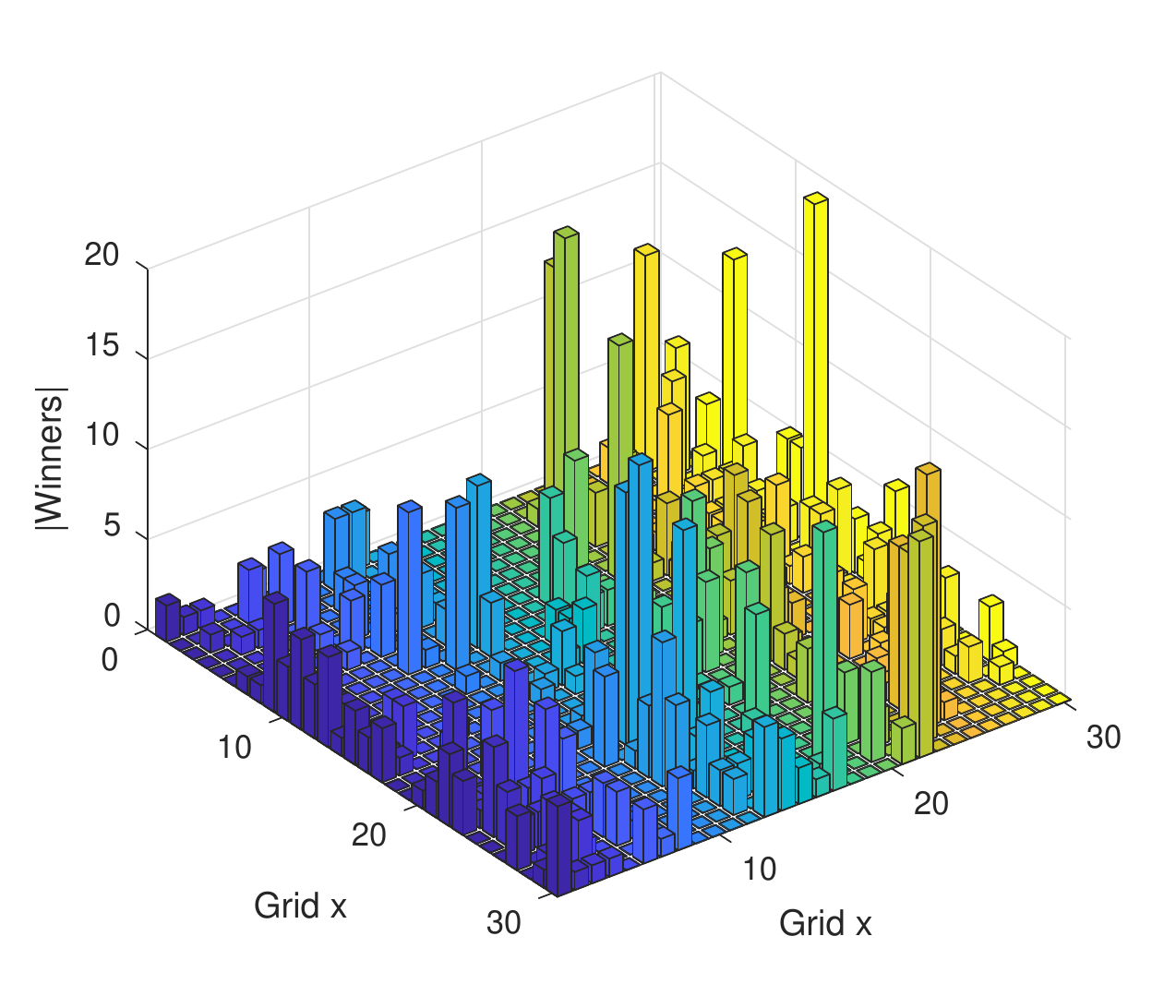}                   }
    \caption{Distribution of winning nodes over several iterations for ECGFiveDays data set: (a) Nodes activated in the initial iteration. (b) Nodes activated in the fifth iteration. (c) Nodes activated in the final (10$^{th}$) iteration. }    
    \label{f:iterations}
    \end{figure*}

\begin{figure*}
        \centering
            \subfloat [\label{f:ECGFiveDays1}]{%
          \includegraphics[width=0.33\textwidth]{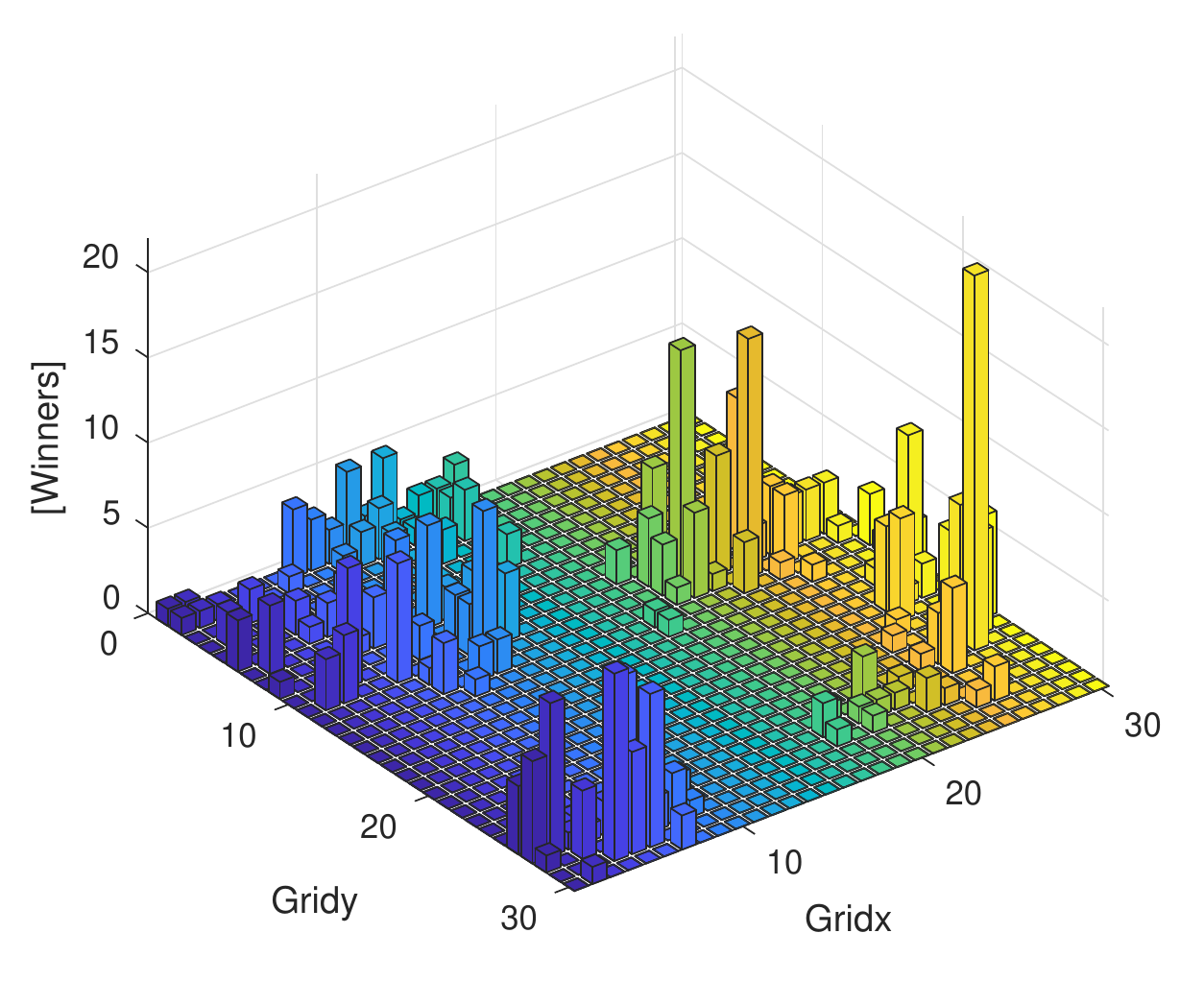}
                   }
             \subfloat[\label{f:ECGFiveDays2}]{%
          \includegraphics[width=0.33\textwidth]{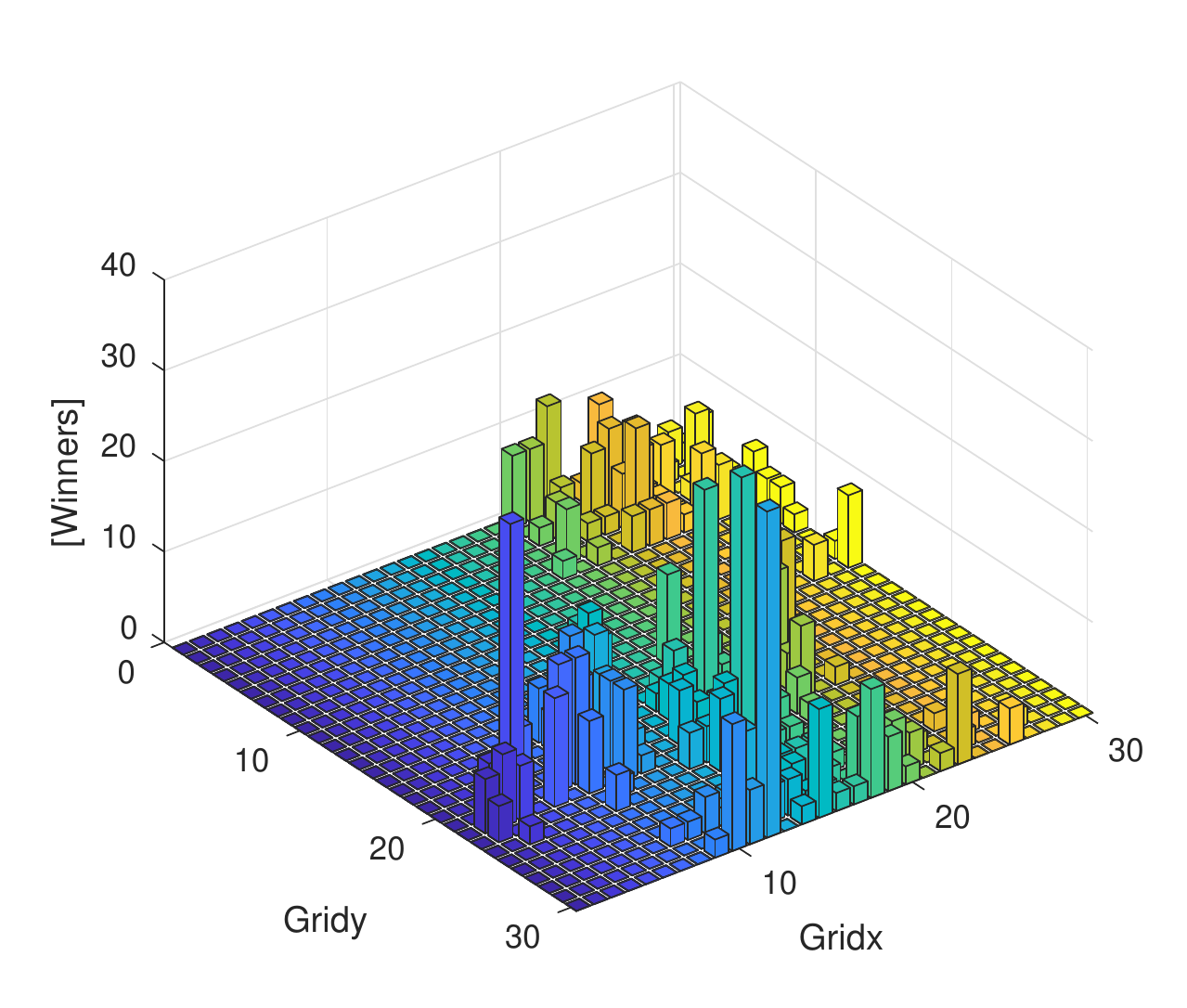}
                   }
            \subfloat[\label{f:ECGFiveDays_tsne}]{%
          \includegraphics[width=0.33\textwidth]{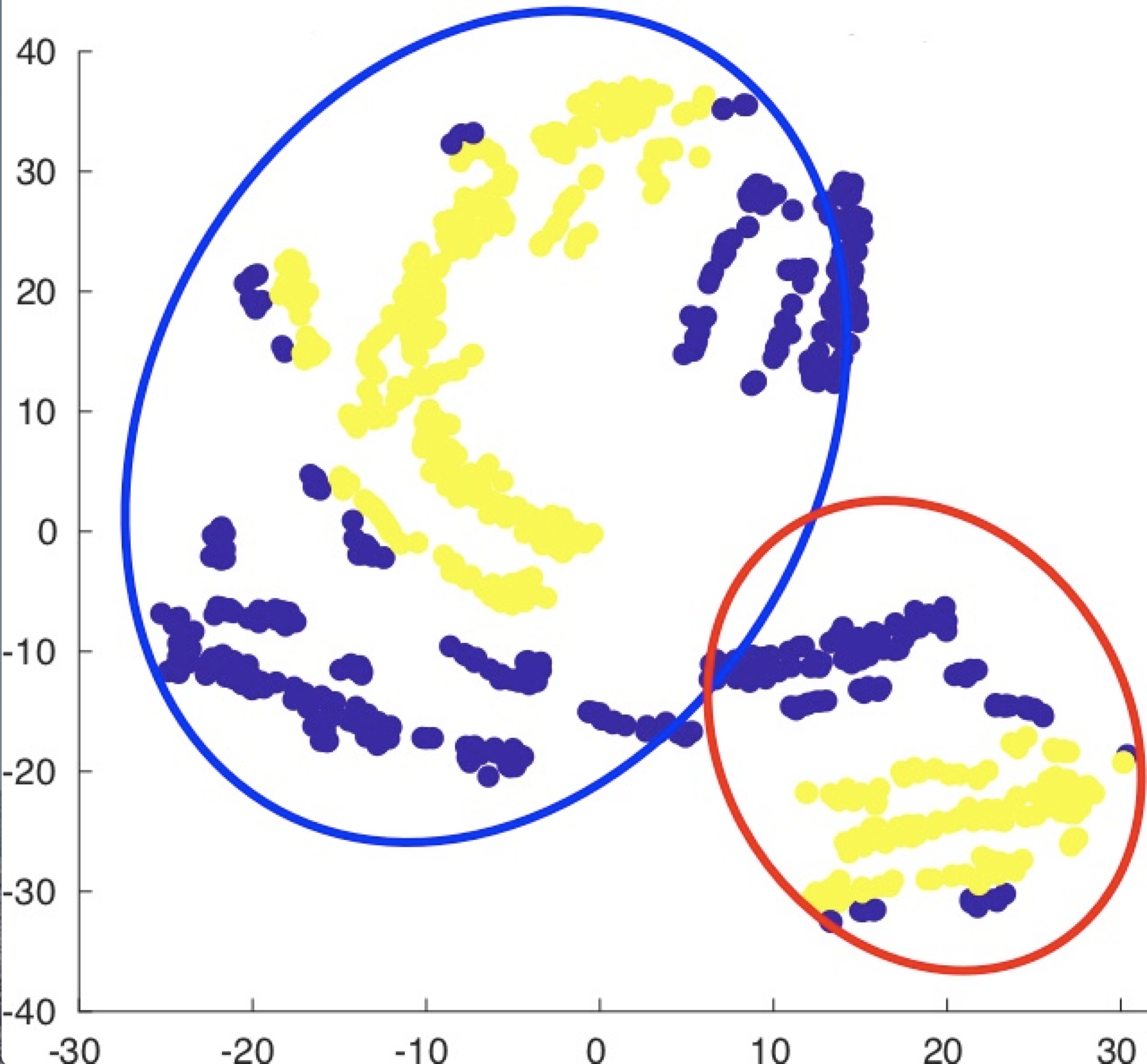}
                   }
    \caption{The distribution of the classes according to the SOEM for ECGFiveDays data set (a) class 1, (b) class 2 and (c) via t-SNE (class 1 in blue, class 2 in yellow). The blue and red circles show clusters selected by kmeans.}
                 \label{f:UCR}
    \end{figure*}

Figure~\ref{f:iterations} shows the output in one of these time series sets; the 'ECGFiveDays' data set. This contains 884 times series, each of length $N=136$ which fall into only $K=2$ classes, 442 series in each class, and so is ideal for visualisation. Figure~\ref{f:it_ex_1} shows the the $30 \times 30$ grid with the Z-axis giving the number of times that node was the winning node. After iteration 1 it can be seen that the time series are uniformly distributed across the grid. After 5 iterations the distribution becomes more concentrated with several clusters emerging (Figure~\ref{f:it_ex_2}). By iteration 10 (Figure~\ref{f:it_ex_3}) the time series have been distributed into clusters in the grid; although this is clearer if we examine the known classes. 
Figures~\ref{f:UCR}a,b split Figure~\ref{f:it_ex_3} into the locations for the 2 known classes. As can be seen class 1 is distributed into 4 clusters with clear gaps between the clusters while class 2 occupies two clusters. Note also that while the clusters are close there is no overlap demonstrating that the SOEM has been successful clustering the time series in this case. Figure~\ref{f:ECGFiveDays_tsne} shows a projection of the 884 time series into 2-D using t-SNE. This shows again that the two classes do occupy distinct locations but in multiple sub clusters (there may be sub-classes unknown during labelling). We applied k-means to the projections in Figure~\ref{f:ECGFiveDays_tsne} but it did not perform well. The average silhouette width indicates the number of clusters is estimated to be between 4 to 8.
    
\begin{figure*}
           \subfloat[\label{f:UCR_db_index}]{%
      \includegraphics[width=.5\textwidth]{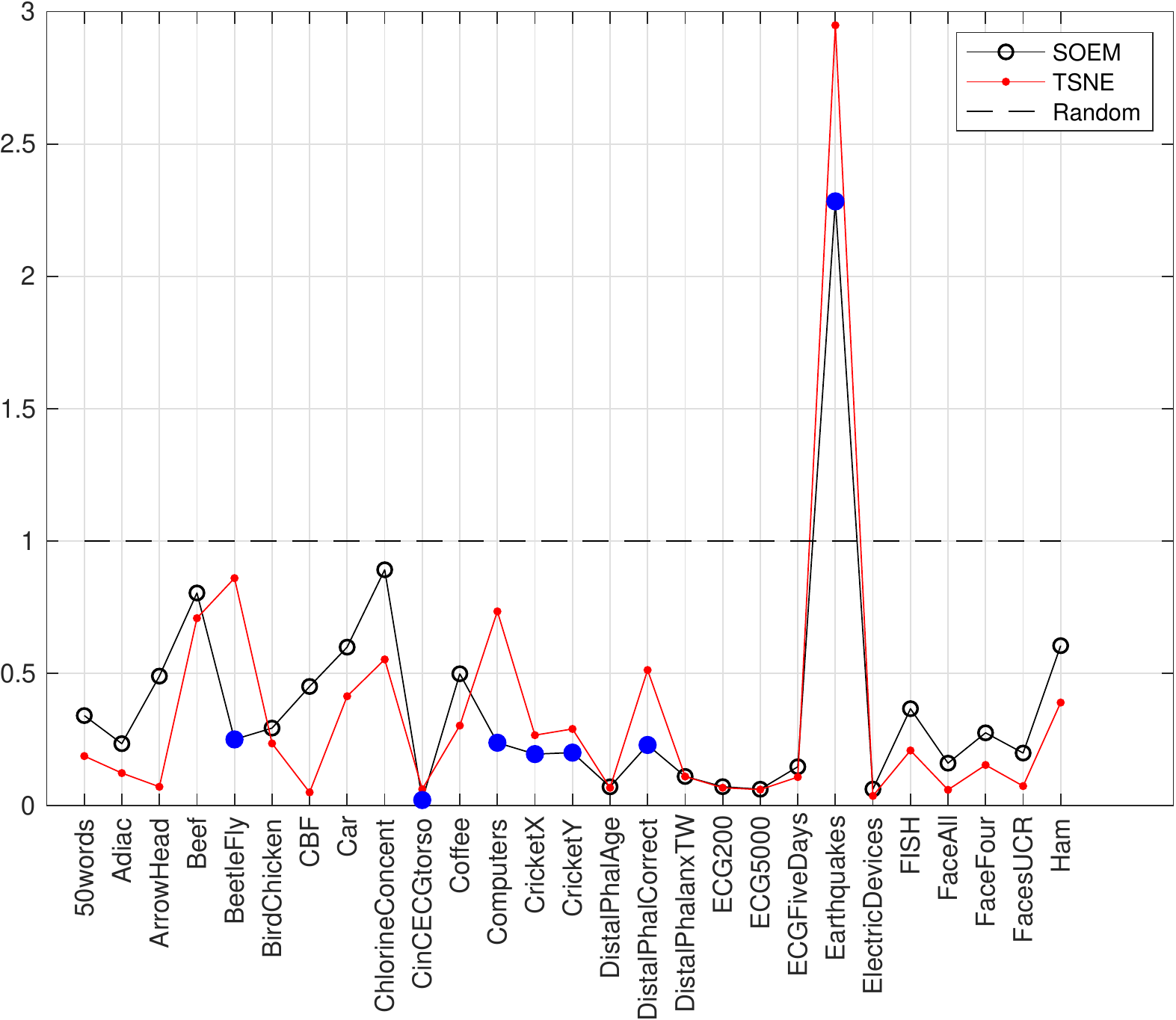}
               }
              \subfloat[\label{f:UCR_pca}]{%
      \includegraphics[width=.5\textwidth]{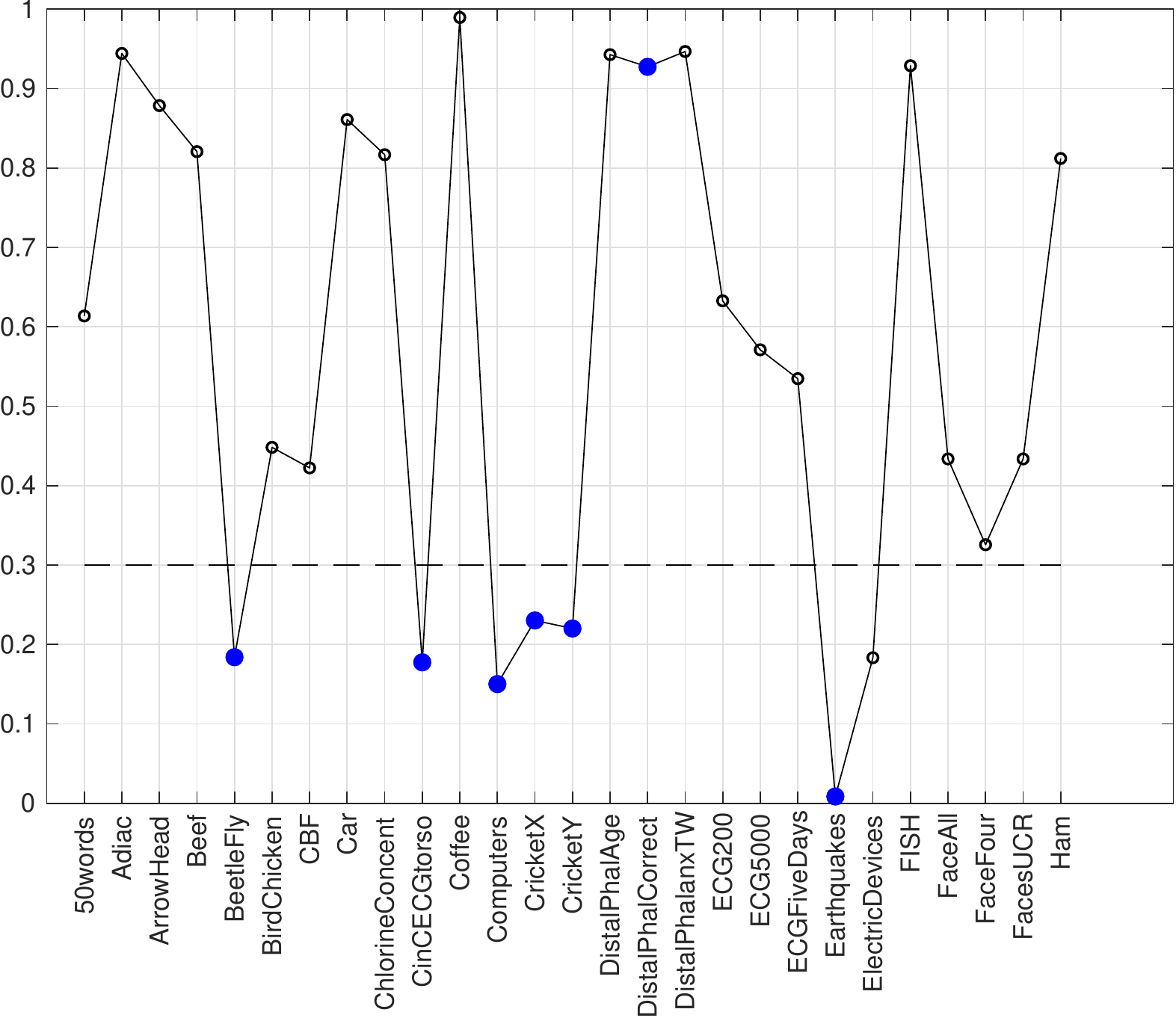}
               }        
\caption{Results on 27 UCR time series: (a) Ratio of DB index for SOEM and t-SNE to a randomly assigned map (black line and red line respectively.) (b) Percentage of variance explained by the first principal component.}
             \label{f:UCR_accu}
\end{figure*}

To summarise the performance of the SOEM across all 27 data sets in the collection we compare the DB index achieved with that from t-SNE. In order to get a reference point we normalise the DB relative to the DB obtained using a uniformly distributed random assignment of the class locations in a square grid (Figure \ref{f:UCR_db_index}). As can be seen both the SOEM and t-SNE perform significantly better than random in all but one case (the earthquake data set). For most cases t-SNE is superior to the SOEM but there are several exceptions shown with blue dots in Figure~\ref{f:UCR_db_index}. On further investigation it was found that the SOEM performs better than t-SNE when the time series are not aligned. This is demonstrated by examining the percentage of variance explained by the first component in a PCA projection of the time series data (Figure~\ref{f:UCR_pca}) (the blue dots in Figures~\ref{f:UCR_accu} (a) and (b) correspond).\footnote{When there is a high degree of correlation between time series PCA will require few components to explain most of the variance. Thus we use the first component as a proxy measure of alignment.}

\subsection{Articulary word recognition time series: Multivariate case}
\begin{figure*}
        \subfloat[\label{mv_1a}]{%
      \includegraphics[width=0.33\textwidth]{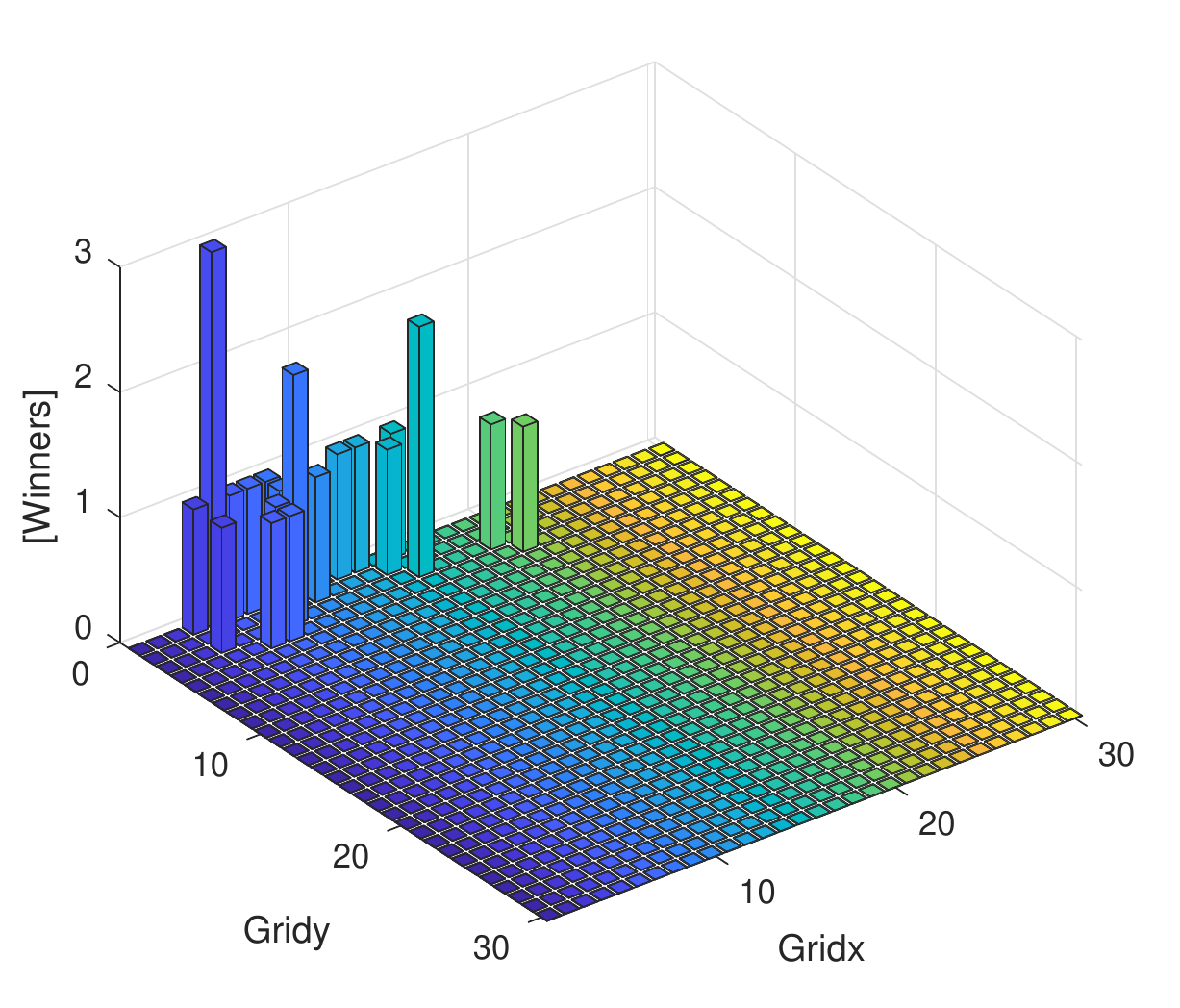}
               }
         \subfloat[\label{mv_1b}]{%
      \includegraphics[width=0.33\textwidth]{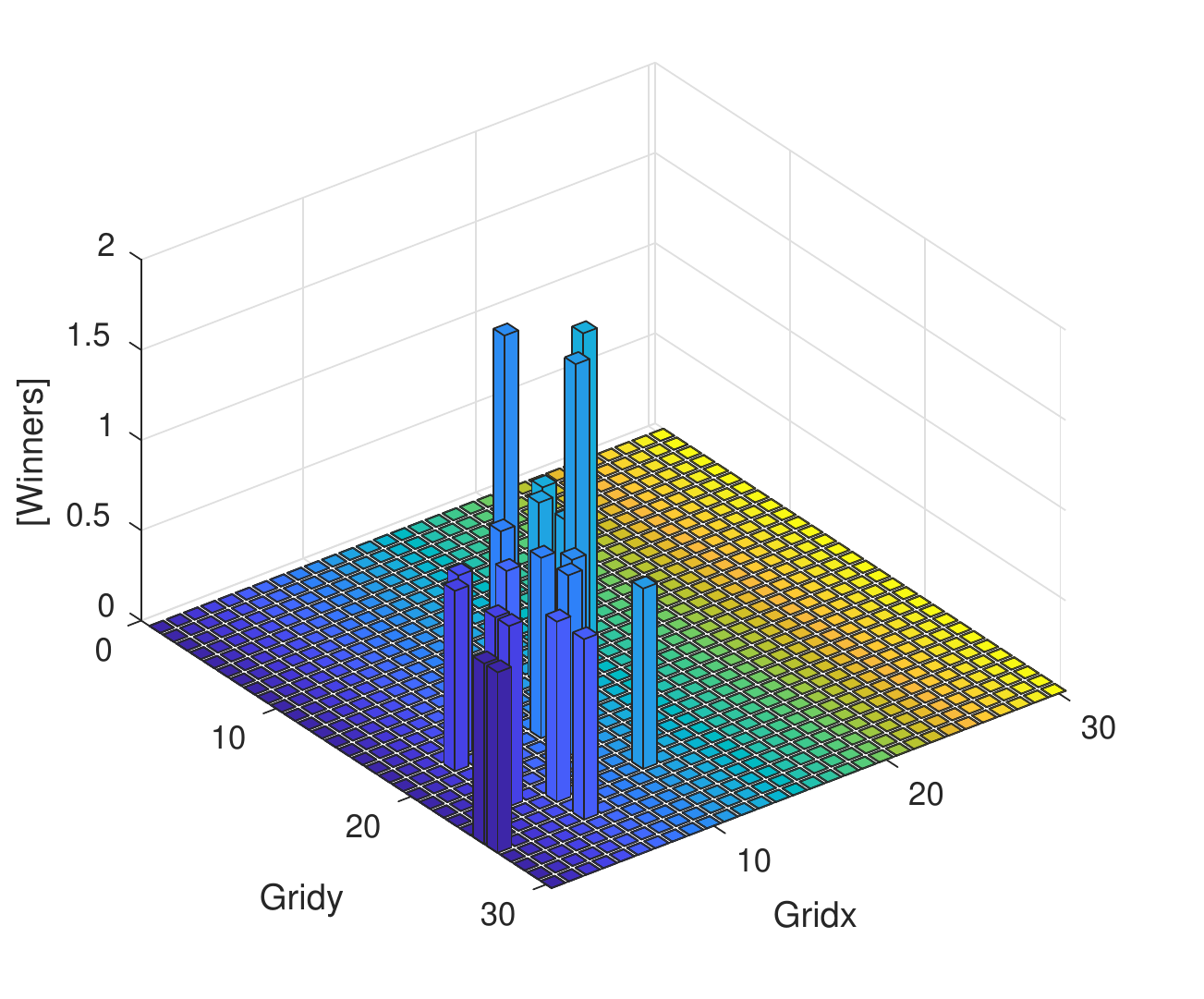}
               }
            \subfloat[\label{mv_1c}]{%
      \includegraphics[width=0.33\textwidth]{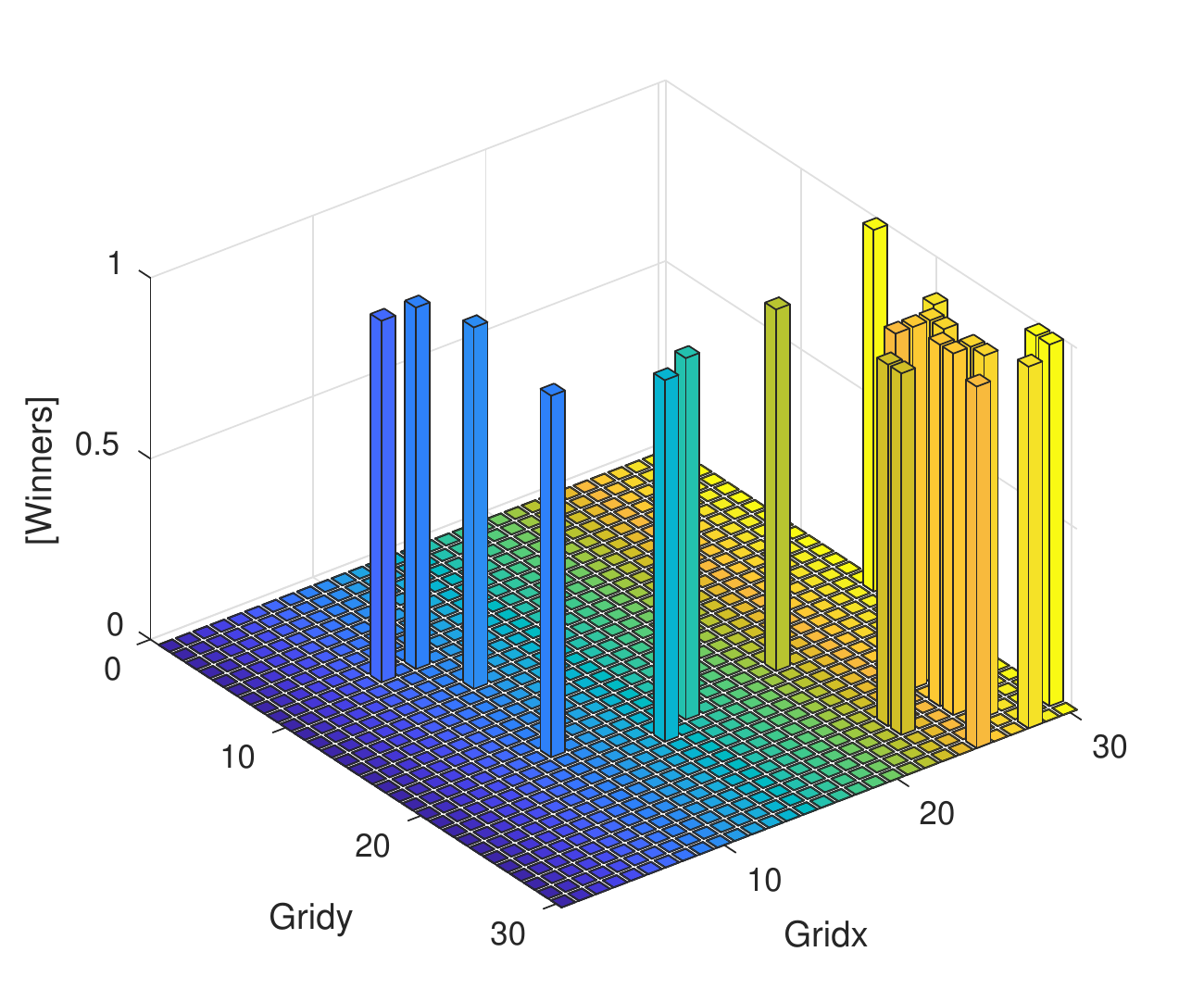}
               }
               
              \caption{The distribution of the classes according to the SOEM for multivariate articulary word recognition data set (a) class 1, (b) class 2, and (c) class 3.}
             \label{f:mv_1}
\end{figure*}

In this Section we show how the SOEM can easily be adapted to clustering in a multivariate data set setting. The Articulary word recognition data set consists of 575 experiments each of length 144 time steps that record the movement of the tongue and lips using 9 sensors. 25 different words (classes) are articulated by each of the 23 participants ($23\times25=575$). Thus we have a $575\times144 \times 9$ data tensor. The embedding dimension was chosen to cover 1/3 of the time interval, $L=48$. The SOEM may be easily extended to the multivariate case using the embedded covariance matrix used in MSSA. Specifically,  Equation~\ref{e:mssa} is used to stack each of the trajectory matrices and the covariance matrix is then created as $\hat{\mathbf{X}}_H \hat{\mathbf{X}}_H ^T$ resulting in 575 $L \times L$ matrices for input to the SOEM. Figure~\ref{f:mv_1} shows the resulting nodes activated for classes 1, 2, and 3. As can be seen the SOEM results in compact clusters for these classes and for most of the others with some exceptions. 

\begin{figure*}
        \subfloat[\label{f:mv_2a}]{%
      \includegraphics[width=0.5\textwidth]{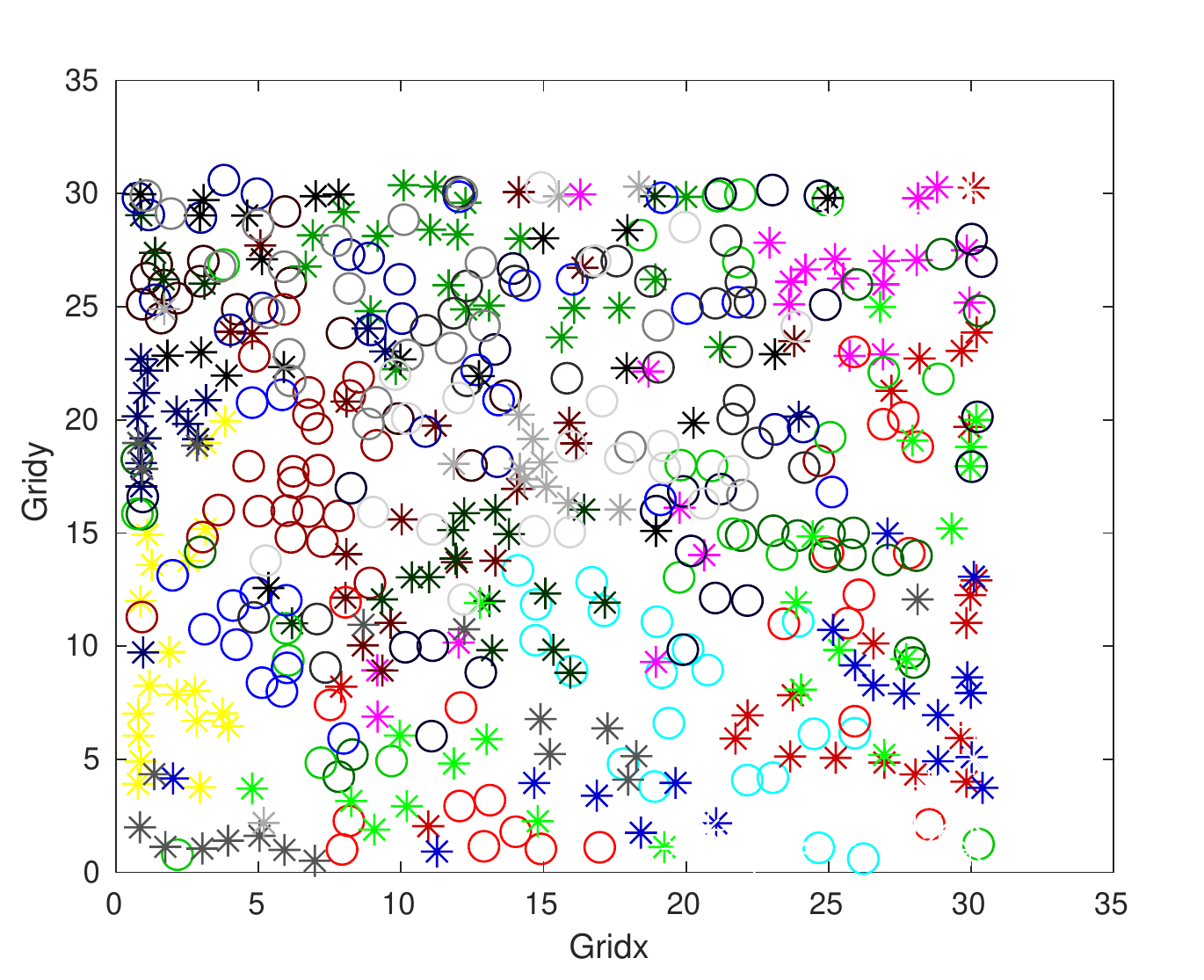}
               }
         \subfloat[\label{f:mv_2b}]{%
      \includegraphics[width=0.5\textwidth]{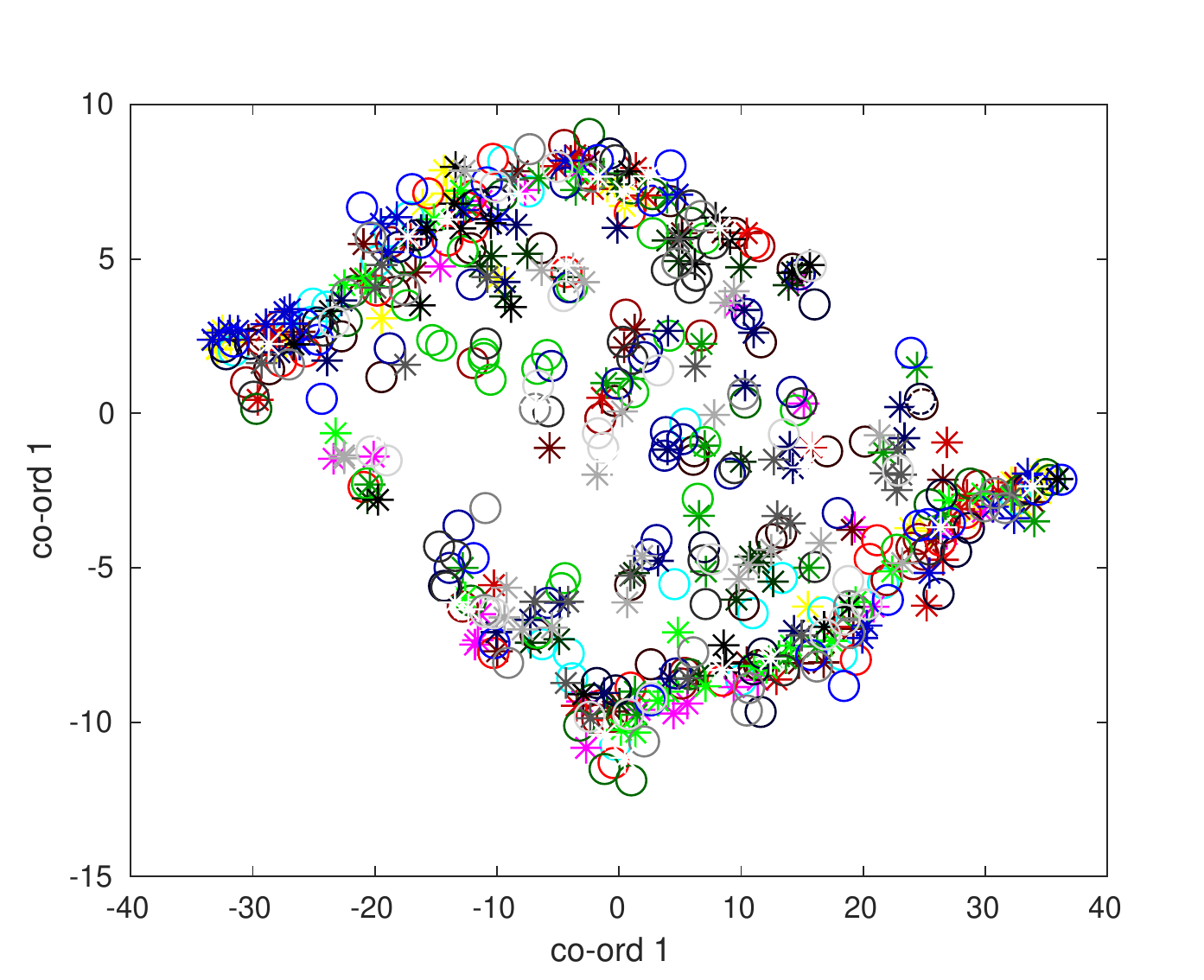}
               }

              \caption{2-D projections for all classes of multivariate articulary word recognition data set using (a) SOEM (note we add a small amount of noise to the grid locations so as to allow visualisation of overlapping classes), and (b) tSNE. (Best viewed in colour.)}
             \label{f:mv_2}
\end{figure*}

In order to compare the results we applied tSNE to the data twice. In the first round each experiment is reduced from a $144 \times 9$ matrix to a $144 \times 1$ using tSNE. These are then appended giving a $575\times144$ matrix which is then reduced to 2-D using a second round of tSNE. The resulting clusters are shown in Figure~\ref{f:mv_2b}. In comparison with Figure~\ref{f:mv_2a} we see that tSNE has done badly in clustering this multivariate data. The DB index for the SOEM, tSNE and a random assignment are 2.53, 15.89, and 11.31, respectively. Thus the SOEM performs significantly better than the alternatives examined. 

\subsection{USA unemployment rate series}\label{s:usa}
The previous experiments involved clustering data and validating against known classes. In this section we examine the forecasting ability (classes are unknown). Specifically, clustering is a pre-processing step MSSA is then applied to the clusters and we examine the subsequent forecast performance. As MSSA requires matching components in the time series this experiment examines how the competing clustering algorithms create clusters with matching components.

The data used here is the mainland USA monthly unemployment rates from Jan 1976 to Dec 2013 ($456$ months $\times 48$ states). The first 304 points are used for training and the remaining 1/3 are used as a hold-out set for forecast performance evaluation. $L$ is set to 39 which is sufficient to account for 3 seasons (years) and covers approximately 1/10$^{th}$ of the data set. 

\begin{figure*} 
        \subfloat[\label{f:iteration_1}]{%
      \includegraphics[width=0.33\textwidth]{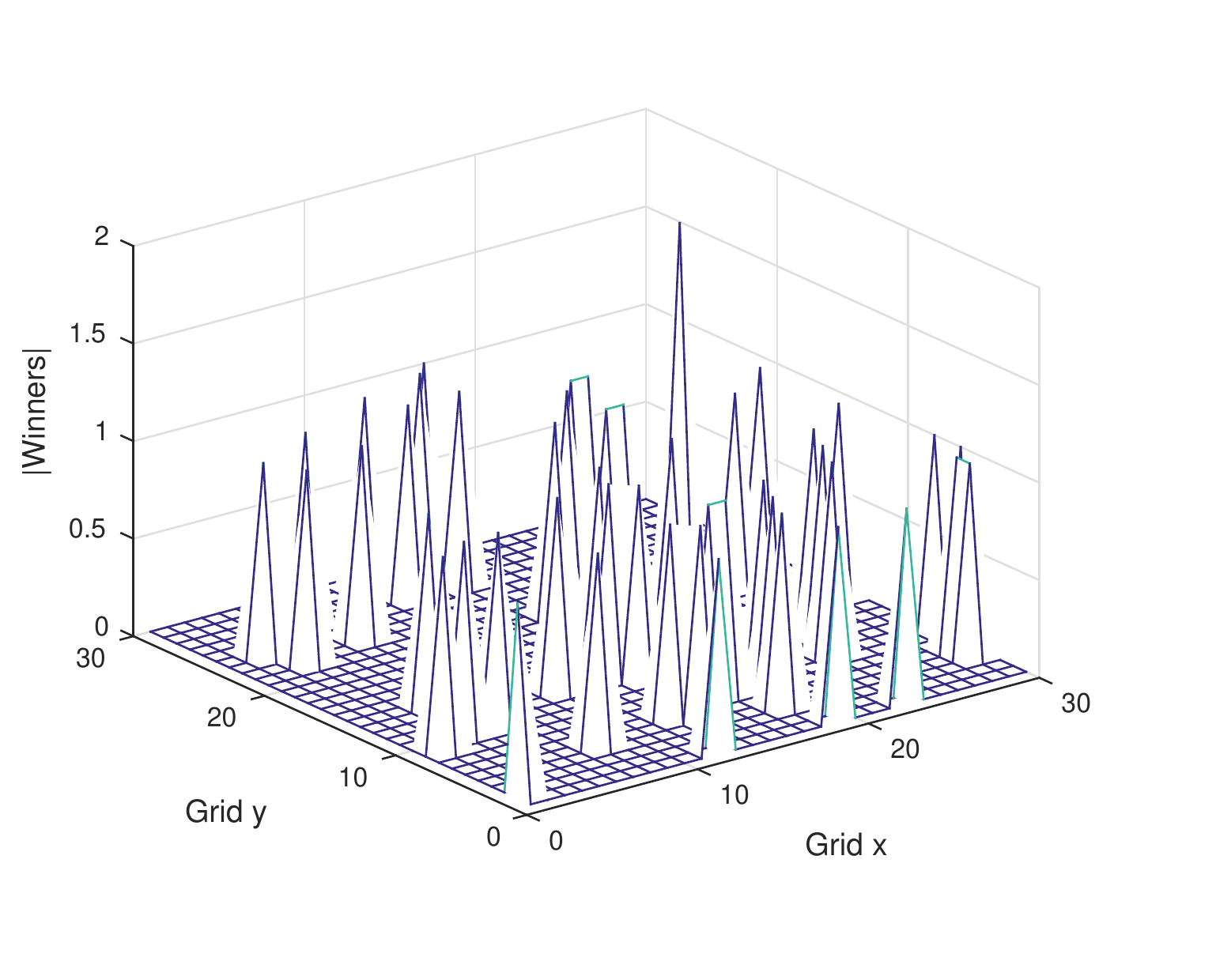}
               }
         \subfloat[\label{f:iteration_2}]{%
      \includegraphics[width=0.33\textwidth]{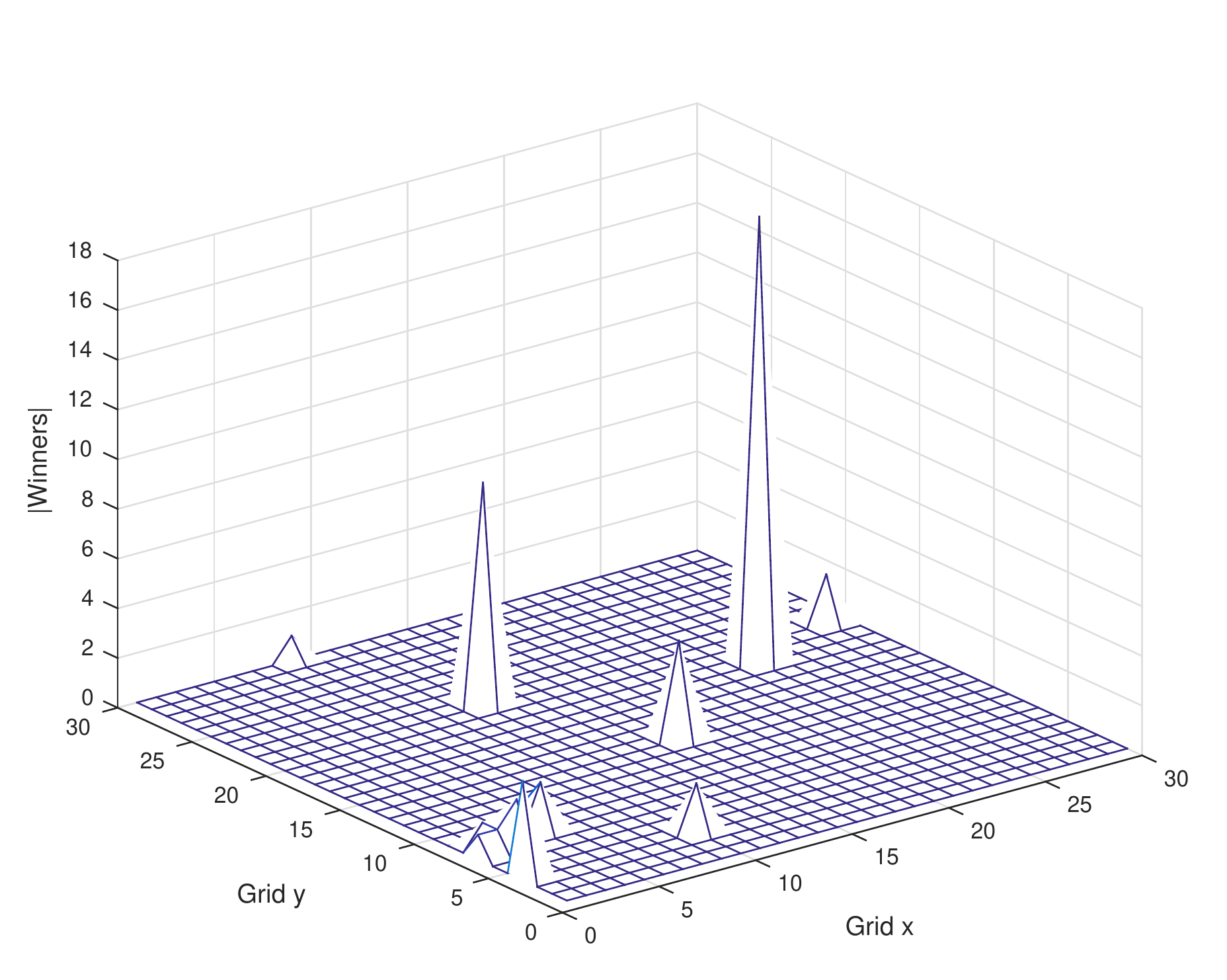}
             }
               \centering
        \subfloat[\label{f:iteration_final}]{%
      \includegraphics[width=0.33\textwidth]{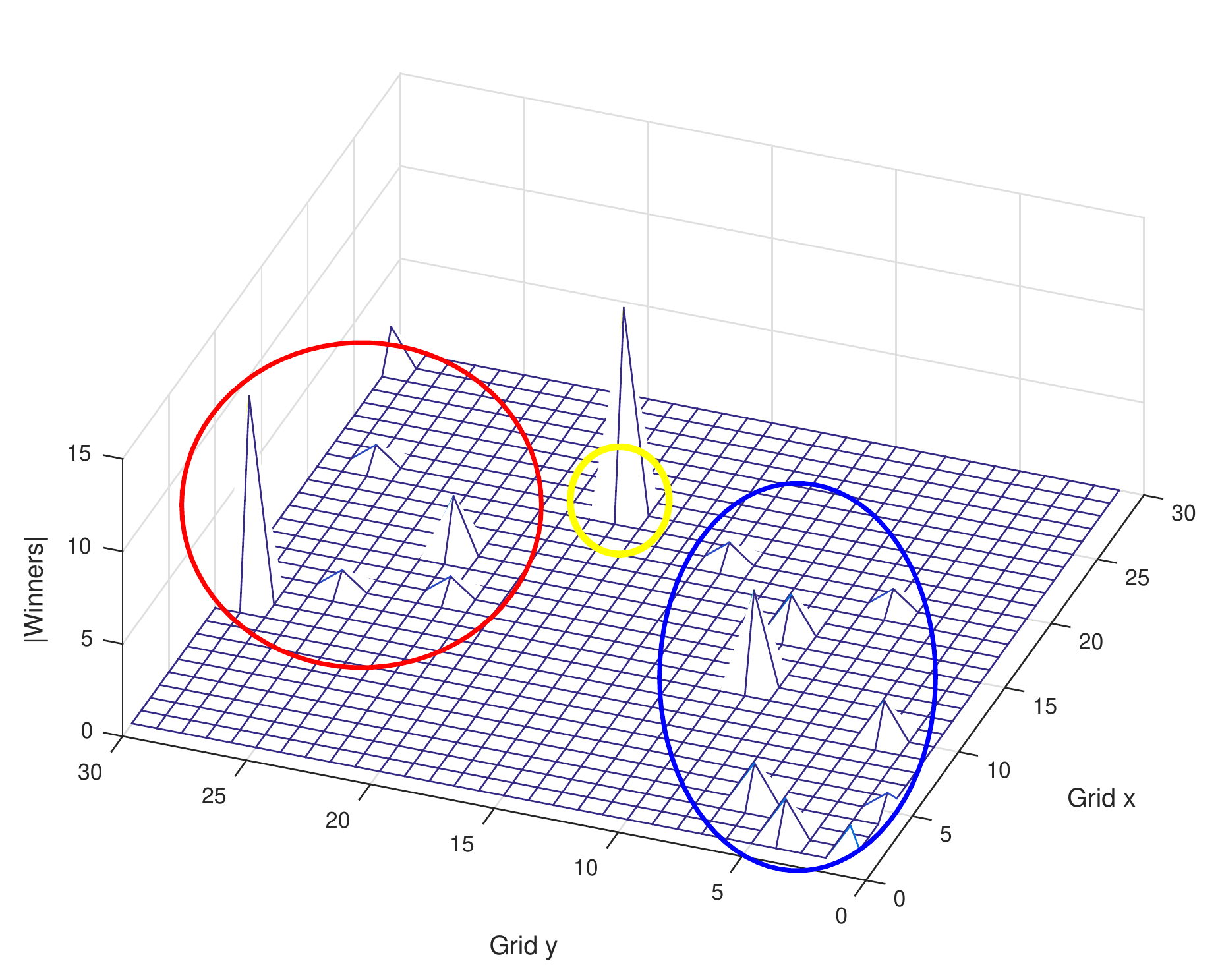}
               }
\caption{Distribution of winning nodes over several iterations for USA unemployment rate series: (a) Nodes activated in the initial iteration. (b) Nodes activated in the second iteration. (c) Nodes activated in the final (10$^{th}$) iteration.
}
\label{f:distance_USA}
\end{figure*}

Figure~\ref{f:distance_USA} shows how the SOEM progresses from the the first iteration (uniformly distributed), to the second iteration where clustering starts to appear and the final iteration where 3 distinct groups are evident; the red cluster on the left, the blue cluster and the yellow cluster in the middle which borders both larger clusters.
\subsubsection{Topographic accuracy}
In this Section we examine the topological ordering given by the SOEM (see ~\cite{tatoian2016self} for a discussion on how to validate the SOFM). Two states are selected from the blue cluster (California and New York) and two states from the red cluster (Iowa and Texas). Specifically we examine the deviations minimised in Equation (\ref{e:soem_opt}), i.e. $\off_2 (||U_{i,j},C||)$ $\forall \{i,j\}$, to see if the deviation increases with distance from the winning node (as evidence that the SOEM is performing topological ordering \cite{tatoian2016self}). Figure~\ref{f:activations_4_cities} shows how the deviations vary for the four example states. As can be seen California and New York have a lower value towards grid location (0,0) (the blue cluster) while Texas and Iowa prefer location (30,30). The key point in these figures is that the transition across the grid is smooth demonstrating the topological ordering for these points. 

\begin{figure*}
        \subfloat[\label{f:California}]{%
      \includegraphics[width=0.5\textwidth]{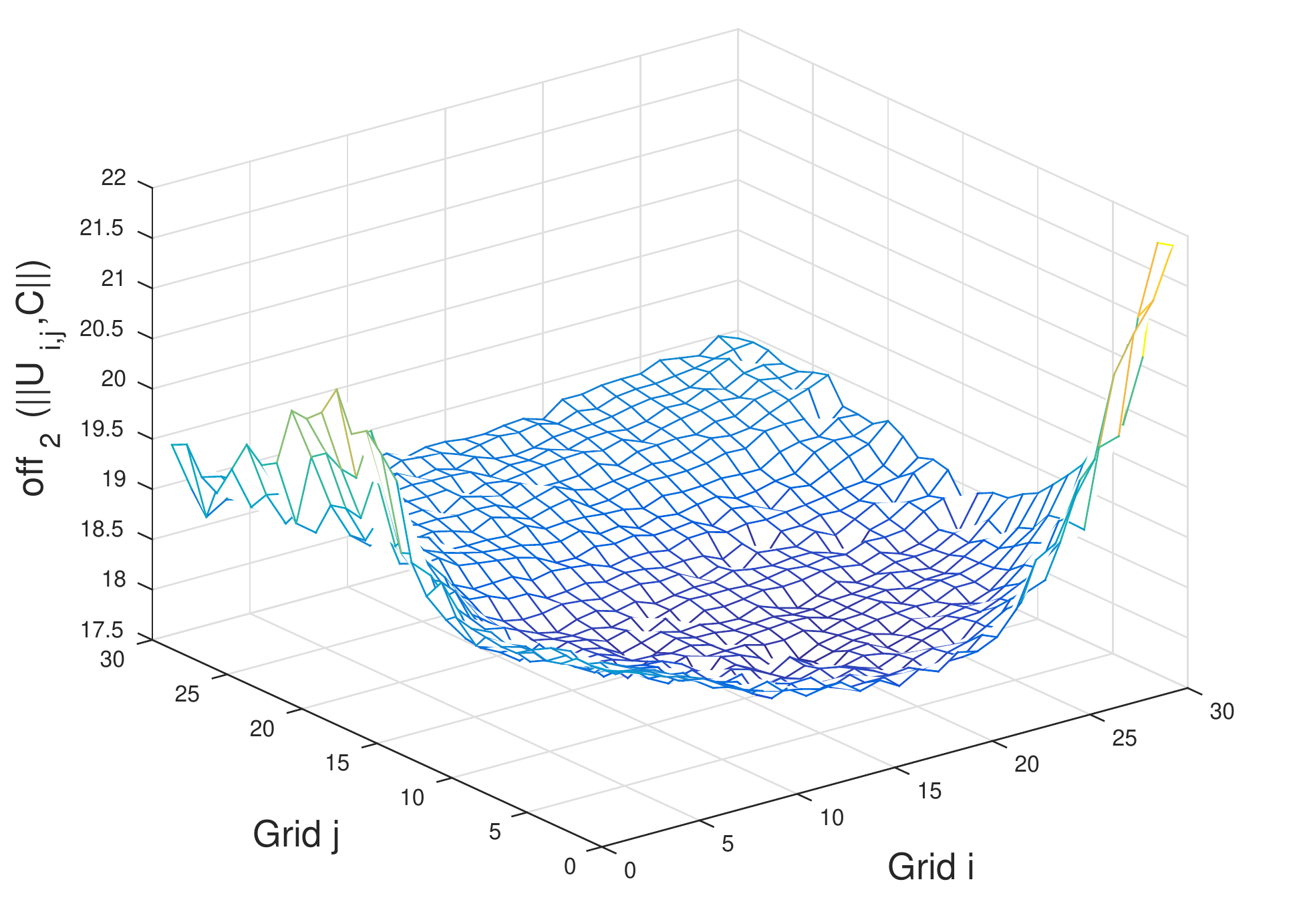}
               }
         \subfloat[\label{f:new_york}]{%
      \includegraphics[width=0.5\textwidth]{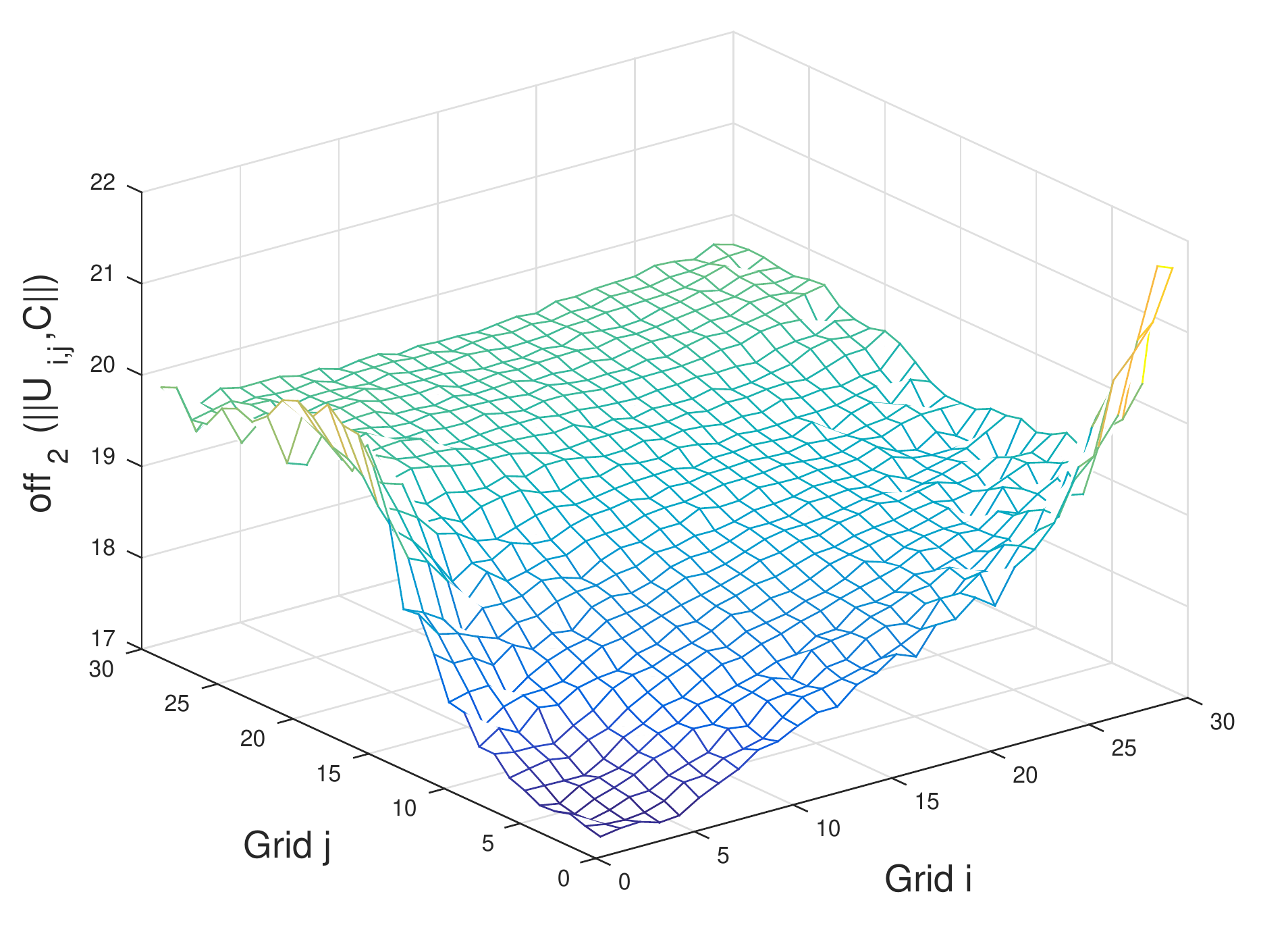}
               }
              
        \subfloat[\label{f:texas}]{%
      \includegraphics[width=0.5\textwidth]{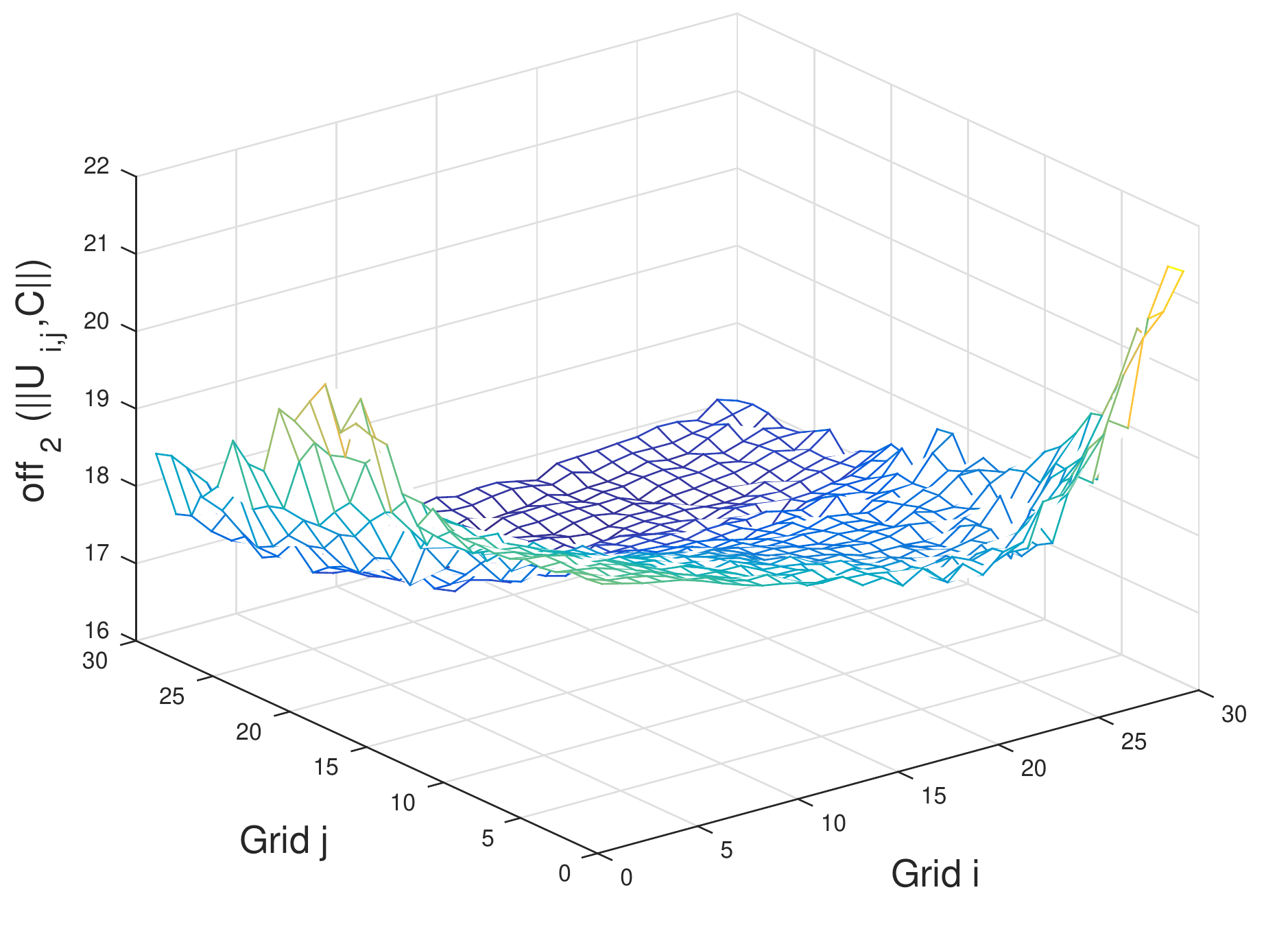}
               }
         \subfloat[\label{f:iowa}]{%
      \includegraphics[width=0.5\textwidth]{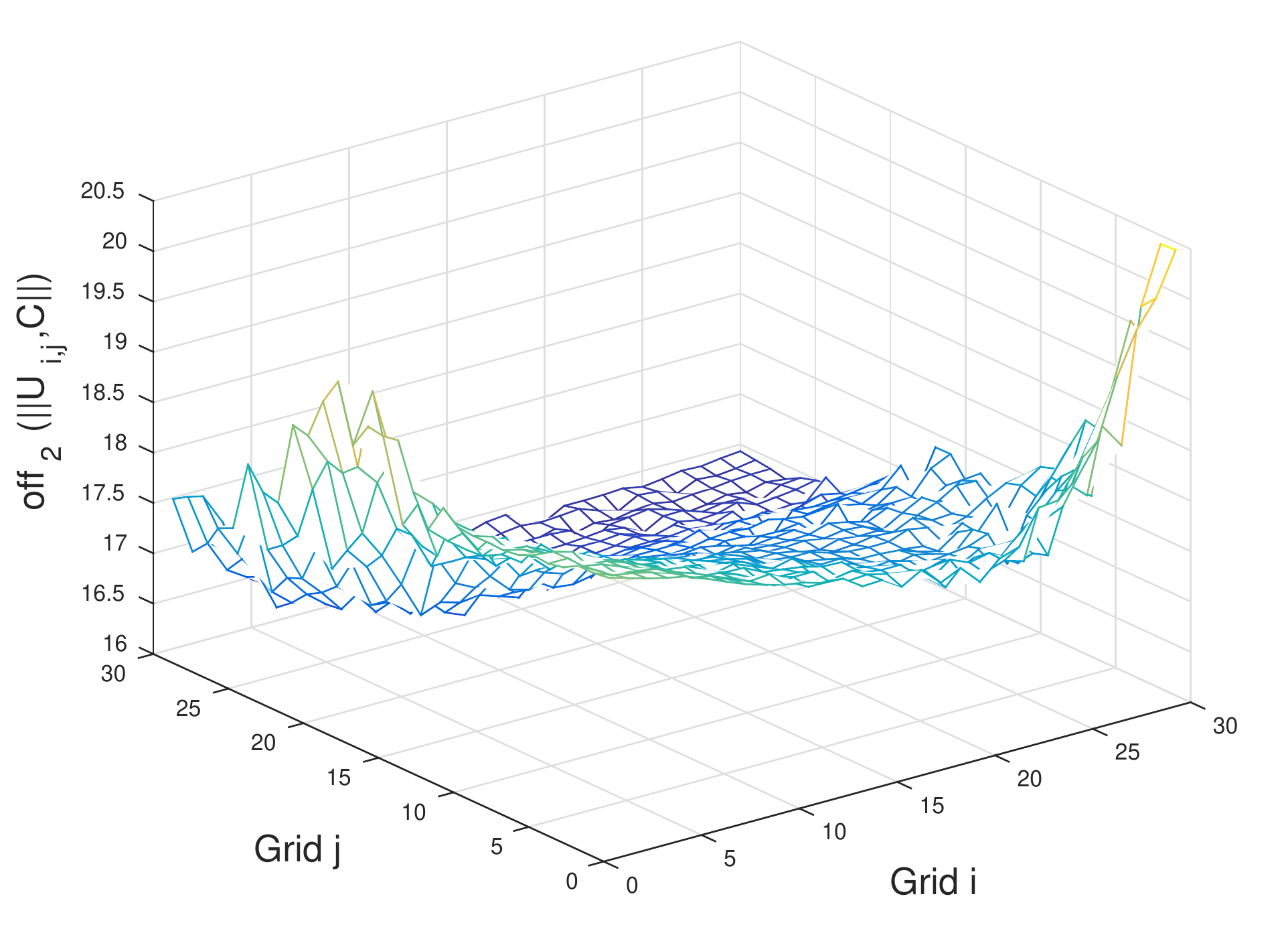}
             }
             \caption{Distance measure,  $\off_2 (||U_{i,j},C||)$, for four sample American states. (a) California. (b) New York (c) Texas (d) Iowa.}
             \label{f:activations_4_ities}
\end{figure*}

To evaluate the convergence of a Kohonen neural network Tatoian  and  Hamel~\cite{tatoian2016self} propose two measures. The first measure is the \emph{topographic accuracy} defined as the average number of neighbours of winning nodes that came second:
 \begin{equation} \label{e:topo_error}
  \epsilon_{\mathcal{T}} = 1/n \sum_{i=1}^{n} 1/||s^{(i)},s^{(i)} _2||_\mathcal{H} 
 \end{equation}
where $||s^{(i)},s^{(i)} _2||_\mathcal{H}$ is the distance between the location of the winning node $s^{(i)}$, for input $i$ and that location which came second, $s_2^{(i)}$. In addition, $||\cdot ||_\mathcal{H}$, denotes the \emph{Hamming} distance in the sense that if $s^{(i)}$ and $s_2^{(i)}$ are not neighbours $||\cdot ||_\mathcal{H}=0$. In an (ideal) topologically ordered space the winning and second place nodes would always be neighbours and this is the motivation for the measure in~(\ref{e:topo_error}). However, the value of $\epsilon_{\mathcal{T}}$ for the map trained by US unemployment datasets is $0.1875$ showing that approximately $20\%$ of the time the $2^{nd}$ best location is a neighbour of the winning node. However, this number does not concur with the general impression conveyed by Figures~\ref{f:California} to~\ref{f:iowa} because it does not take into account the noise present in an empirical ordering. 

\begin{figure}
    \centering
    \includegraphics[width=0.8\textwidth]{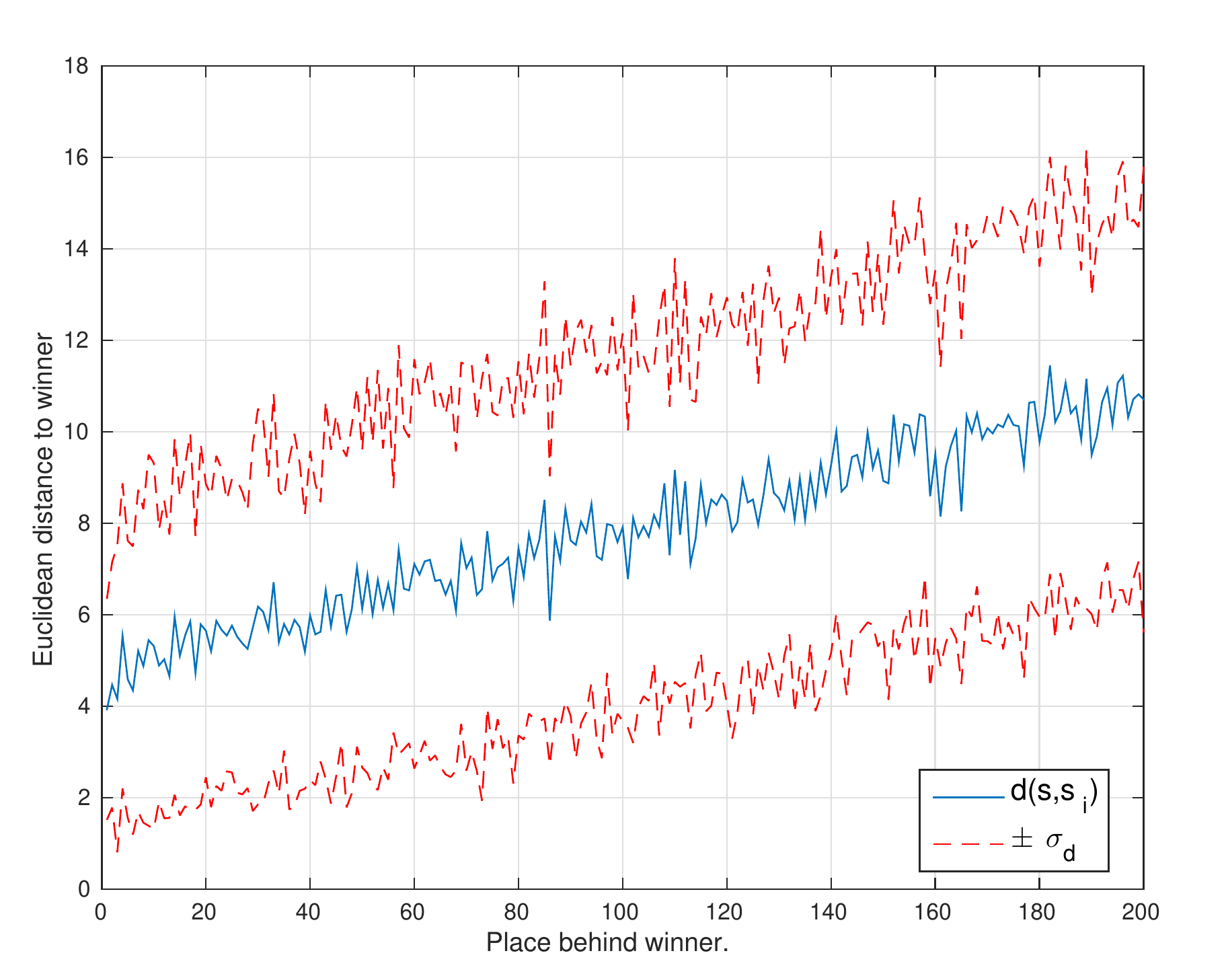}
    \caption{The x-axis is the number of places behind the winning node (i.e. 2$^{nd}$ place, 3rd place etc), the y-axis shows the average Euclidean distance from the winning node of that node.}
    \label{f:koh_top_ordering_errors}
\end{figure}

\label{f:activations_4_cities}
Instead we report the average distance to $s_2$ averaged over all the states which is $3.92$. Indeed, the distance from the $i^{th}$ place loser increases monotonically from the winner as shown in Figure~\ref{f:koh_top_ordering_errors}. This demonstrates that a topological ordering has occurred as there is clear upward trend in the place achieved and the distance to the winner.  
\subsubsection{Comparison with alternate techniques}
In this section we examine the geographical clustering given by several competing techniques. In addition to the clusters obtained from the SOEM we compare t-SNE, JD and Tensor decomposition. For JD based clustering a joint diagonalisation is performed using all 48 time series embedding matrices. The distribution of the deviations is then clustered via a GMM (full details are given in~\cite{Thesis_Donya} and are beyond the scope of this paper). The Tucker transform is used to decompose the $48 \times (L \times L)$ embedding matrices into a $3 \times (L \times L)$ where the resulting scores are clustered using kmeans (we also explored different variants of the tensor transform and hierarchical clustering which may be found~\cite{Thesis_Donya}). With the latter alternative we are able to view the covariance matrices as a multidimensional array, tensor, and used a higher dimensional tensor decomposition, the Tucker transform. In all cases we use $K=3$ clusters so as to perform a  fair comparison between the techniques.

\begin{figure*} 
        \subfloat[\label{f:soem_USA}]{%
      \includegraphics[width=0.5\textwidth]{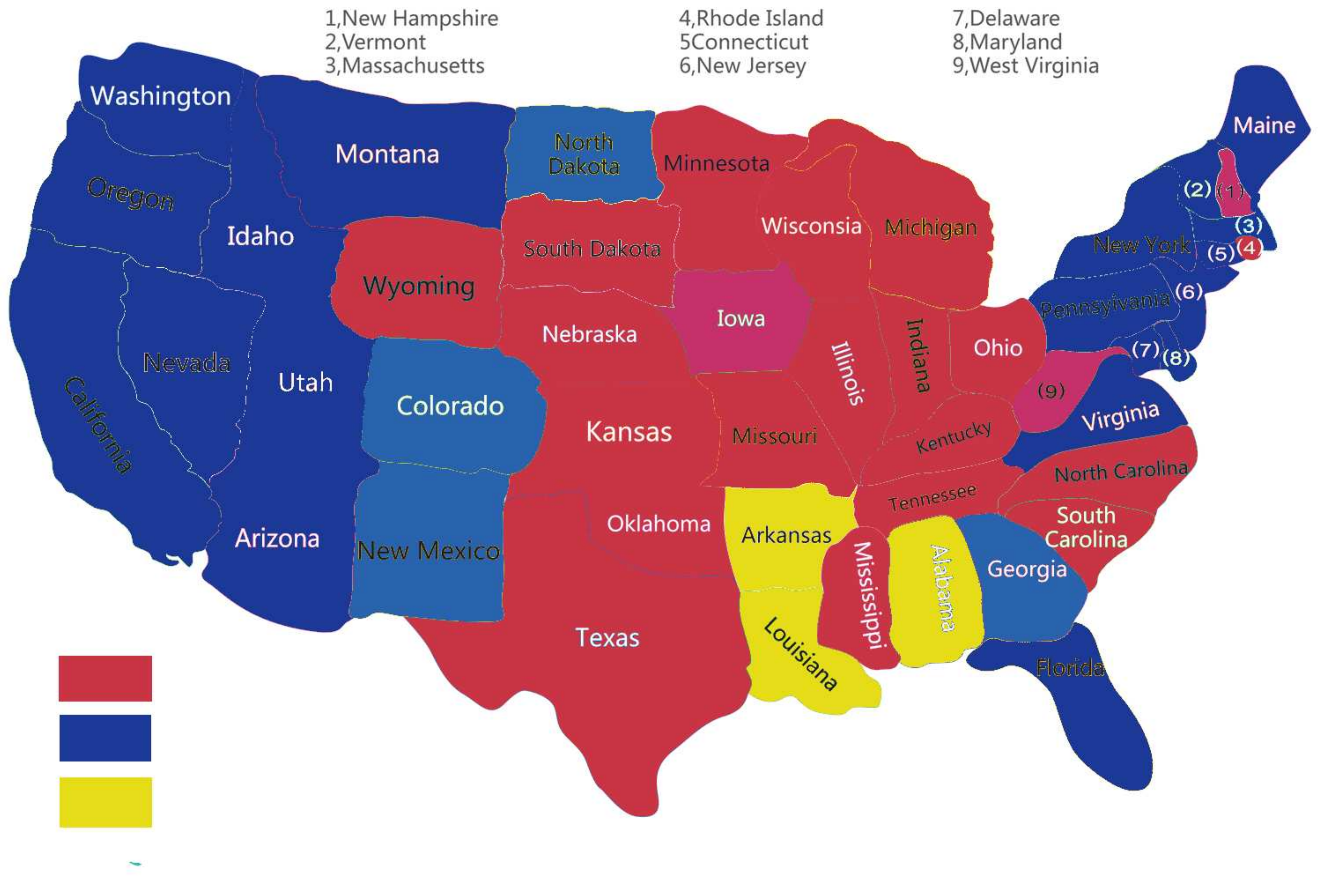}
               }
         \subfloat[\label{f:tsne_USA}]{%
      \includegraphics[width=0.5\textwidth]{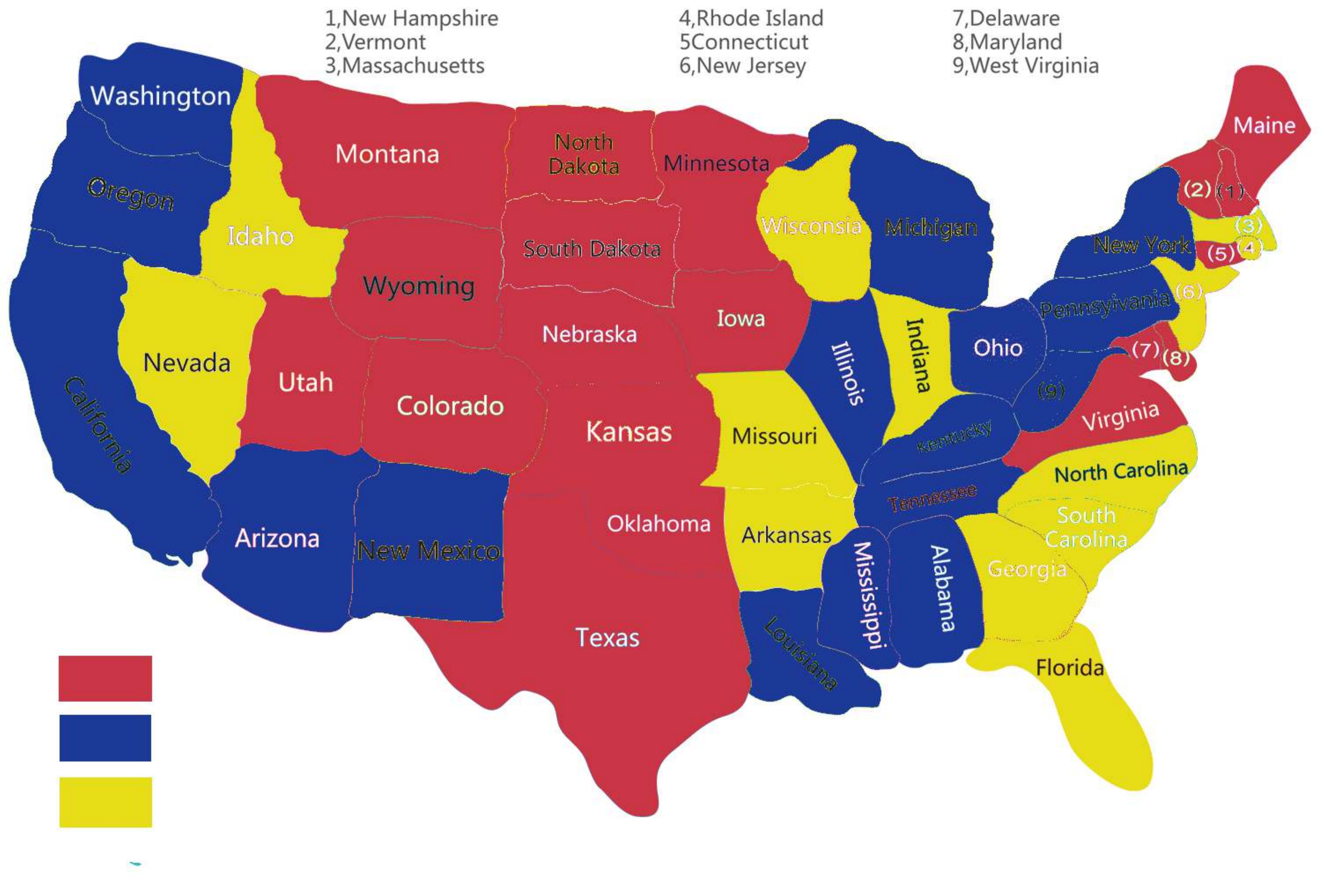}
               }
               
  \subfloat[\label{f:JD_USA}]{%
    \includegraphics[width=0.5\textwidth]{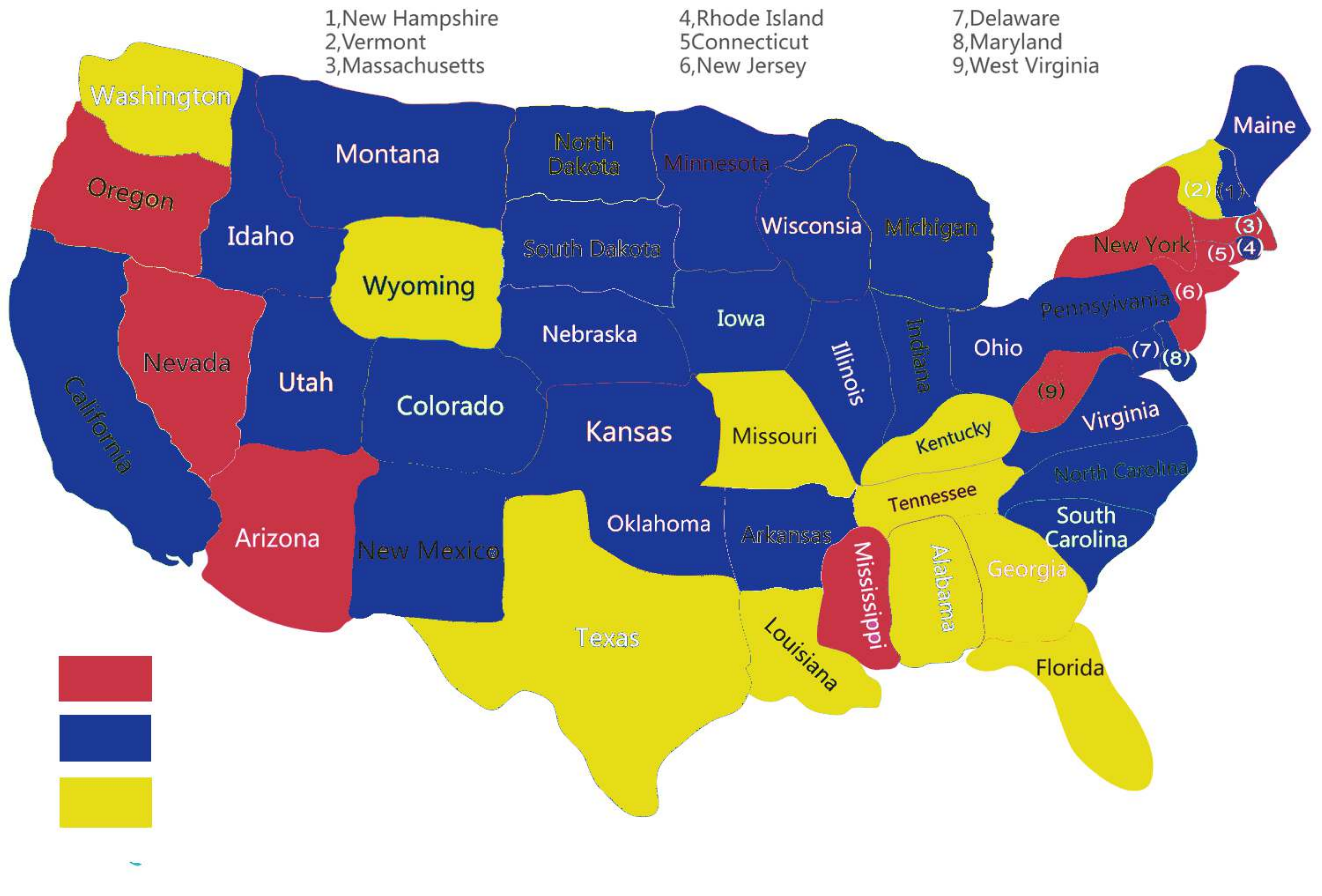}

  }
  \subfloat[\label{f:tensor_USA}]{%
    \includegraphics[width=0.5\textwidth]{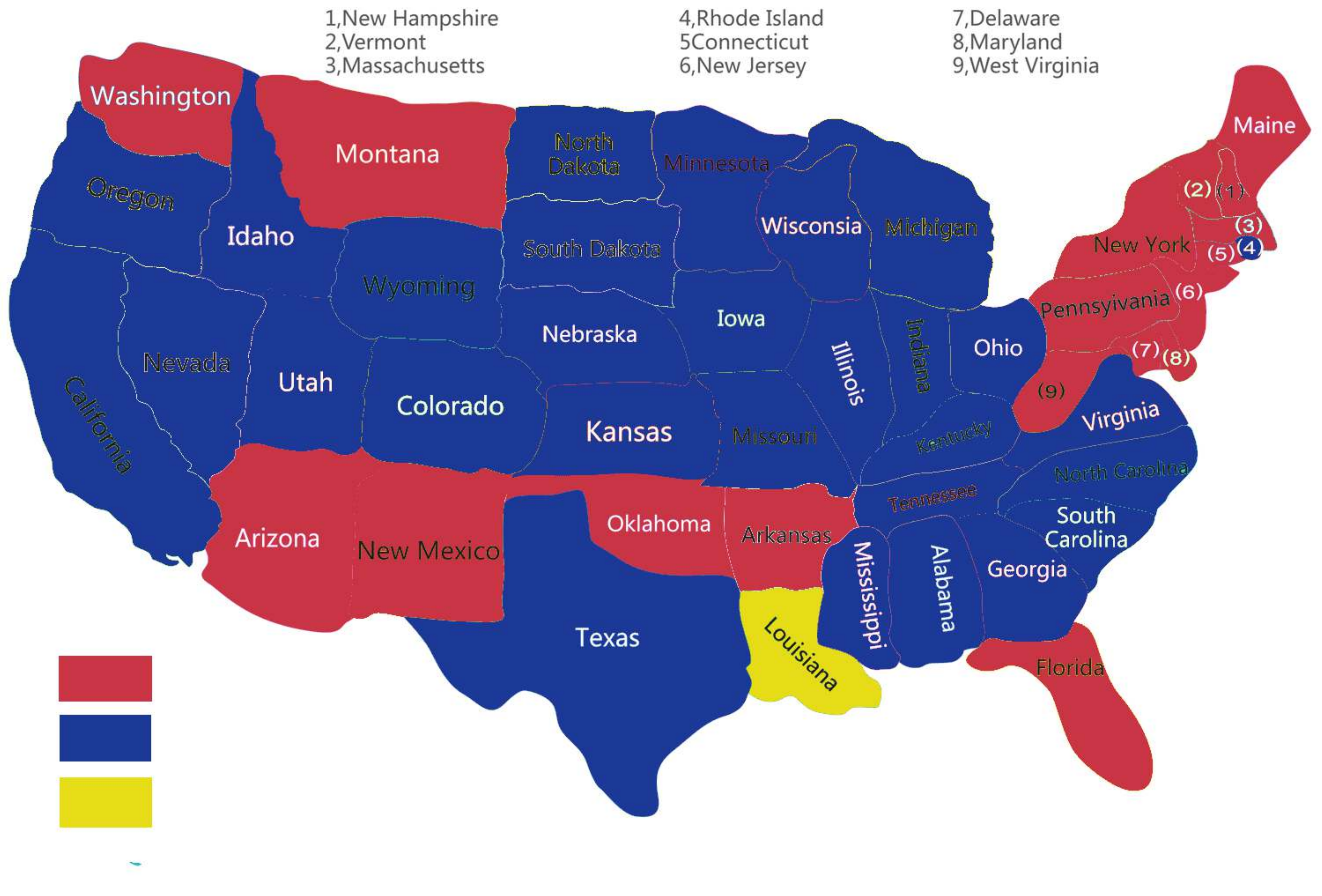}
  }
     \caption{States clustered with (a) SOEM (b) t-SNE (c) JD (d) tensor decomposition}
             \label{f:maps_USA}
\end{figure*}

Figure~\ref{f:maps_USA} presents the clustering of states given by the SOEM, t-SNA, JD and Tensor decomposition based methods. Figure~\ref{f:soem_USA} quite clearly shows a central band of red states and east and west coast states in a blue cluster. In addition, there is also a third yellow cluster of states located in the far South. From an intuitive point of view this clustering appears to make sense as industrial states such as Florida, New York and California are clustered together while the more agricultural central states are also clustered together. Overall there is contiguity in the map; i.e. neighbouring states tend to lie in the same cluster. Figure ~\ref{f:tsne_USA} clusters central states as red states plus Virginia and west coast states in a blue cluster. The yellow cluster of states are located in the far east and mainly surrounded by blue and red clusters. On the other side, states such as New York, Washington and California are clustered together with Ohio, Pennsylvania and Mississippi which leads to a map with less contiguity than in Figure~\ref{f:soem_USA}. According to Figure \ref{f:JD_USA}, 44$\%$ of states in West and Midwest are in one cluster, compared to 22$\%$ in South (and Washington and Wyoming from West). On the other hand, New York, Florida, Nevada and Oregon belong to the same cluster. Again we see a large degree of contiguity between the states but there are also some states clustered together which would not appear to be a natural cluster. For example Vermont is grouped with the southern states and Florida would not typically be clustered with the southern states. Figure \ref{f:tensor_USA} shows the clustering map produced from the tensor decomposition. The results here lead to an unbalanced clusters with most states clustered into one cluster. The analysis above is subjective but informative in that it shows how the competing techniques differ when we examine the geographical differences in clustering. The next section objectively compares their time series prediction performance.

\subsubsection{Clustering and forecasting MSSA}
As mentioned in Section~\ref{s:ssa} MSSA can out-perform univariate SSA when the combined time series have matched components. Thus the aim of clustering is, in this case, to produce three sets of time series, MSSA is then applied to each set to produce forecasts. It is expected that algorithms producing clusters with greater inter-cluster matching will lead to subsequently superior forecasts. The forecasting accuracy is assessed for four different horizons, 1-step ahead, 3 and  6-steps ahead and one year ahead (12-step ahead) (more results and comparisons are presented in ~\cite{Thesis_Donya}).
The overall forecasting results obtained by forecasting via the competing cluster maps are given in Table~\ref{SOEMsum}. The SOEM clusters results in a lower prediction RMSE than the competing techniques in all cases. Surprisingly, the common eigenspace provided by JD had little effect on improving the performance of MSSA forecasting. The reason for that is those common modes between different groups of series are still supporting each other. The Tucker transform as well as t-SNE were found, like JD, to not be an ideal technique to improve forecasting accuracy.  

\begin{table}[!t]
\renewcommand{\arraystretch}{1.5}
\caption{Summary statistics for out-of-sample forecasting accuracy measures for unemployment rate series using MSSA, SOEM-MSSA, JD-MSSA, Tensor-MSSA and t-SNE-MSSA}
\label{SOEMsum}
\centering
\resizebox{\columnwidth}{!}{%
\begin{tabular}{c|ccccc}
\hline
Steps&\multicolumn{5}{c}{RMSE} \\ 
\hline
h&MSSA&SOEM-MSSA&JD-MSSA&Tensor-MSSA&t-SNE-MSSA  \\
\hline
1&  0.12     &   \textbf{0.10} &  0.12  & 0.13  & 0.12 \\
3&  0.37   &   \textbf{0.27} &  0.38  & 0.36  & 0.35 \\
6&  0.77   &   \textbf{0.58} &  0.83  & 0.84  & 0.89 \\
12& 1.68   &   \textbf{1.22} &2.09    & 2.76  & 2.56 \\
\hline
\end{tabular}
}
\end{table}

\section{Conclusion}
A novel technique for clustering time series has been presented, which is based on an adaptation of the Kohonen neural network which is particularly suited to non-aligned time series. In addition, the technique can easily be extended to multivariate time series clustering and to time series which have differing lengths (as long as each time series has length $>L$). As with all clustering techniques success depends on the characteristics in the data that are being targeted. In this case the SOEM targets the auto-covariance structure of the time series and in the case of the results presented in Section~\ref{s:ucr_results} this resulted in better correspondence with the known classes than t-SNE for some time series but not for others. In the forecasting case (Section~\ref{s:usa}), it is perhaps not surprising that the SOEM leads to significantly better performance than the competing techniques as it is targeting the precise characteristic important to MSSA. We also observed in  Section~\ref{s:ucr_results} that several of the known classes were clustered into two well defined clusters perhaps revealing the presence of two sub-classes unknown during the original labelling. As with all clustering techniques we have shown that the SOEM is a useful tool for visualising the inherent distribution of the time series to aid in various tasks such as prediction, classification etc. 

There are several variants of the SOFM which we did not experiment with. For example the size of the grid is fixed while approaches such as those suggested in~\cite{grower} show that growing or shrinking grids can lead to improved clustering. In addition, 3-D SOFM's are popular and there is no reason why our technique cannot be extended to this case.

From a computational complexity viewpoint, creating the embedding matrices and evaluating a trained SOEM have marginal complexity. The core computation is the JD step in Equation(\ref{e:soem_update}) which must be performed $N \times M$ times (where $N,M$ is the size of the grid) each with complexity $\mathcal{O}(L^2)$ leading to an algorithm with complexity of $\mathcal{O}(NML^2)$. A Matlab implementation of our code is correspondingly slow as suggested by the complexity (some simulations take several days to run).\footnote{Sample data and code may be found at github.com/drahmani/SOEM } However, as the calculation performed at each grid location are independent and consist of identical operations, the algorithm should lend itself to a CUDA/GPU implementation with complexity $\mathcal{O}(L^2)$. 

We examined the first measure of an SOFM's performance given by~\cite{tatoian2016self} in Section~\ref{s:usa} and found that the SOEM does indeed provide a topological ordering. The second measure involves showing that the distribution of the input data samples is matched by the distribution of the network. For an SOFM this is easily performed using a t-test \cite{tatoian2016self} as one is comparing two univariate distributions. However, for the SOEM one needs to test if two sets of matrices are sampled from the same matrix distribution. Two sample tests for higher dimensional covariance matrices (for example \cite{li2012}) do exist, although this is one avenue for future research. Indeed, as the SOEM is at an introductory stage it would be interesting to determine when it might converge for general matrices or under what circumstances.  

\section*{Acknowledgment}
This research was supported  by an EPSRC grant EP/N014189/1.

\bibliographystyle{unsrt}  
\bibliography{library}  






\end{document}